\documentclass[10pt]{article} 
\usepackage[preprint]{tmlr}

\usepackage{amsmath,amsfonts,bm}

\def\eqref#1{equation~\ref{#1}}

\def\1{\bm{1}}

\DeclareMathAlphabet{\mathsfit}{\encodingdefault}{\sfdefault}{m}{sl}
\SetMathAlphabet{\mathsfit}{bold}{\encodingdefault}{\sfdefault}{bx}{n}

\usepackage{hyperref}
\usepackage{url}
\usepackage{graphicx}
\usepackage{subcaption}

\title{Bridging Reasoning to Learning: Unmasking Illusions using Complexity Out-of-Distribution Generalization}

\author{\name Mahdi Samiei \email mm.samiei@sharif.edu \\
      \addr Department of Computer Engineering\\
      Sharif University of Technology
      \AND
      \name Arash Marioriyad \email arashmarioriyad@gmail.com \\
      \addr Department of Computer Engineering\\
      Sharif University of Technology
      \AND
      \name Arman Tahmasebi-Zadeh \email arman.tahmasebi345@sharif.edu\\
      \addr Department of Computer Engineering\\
      Sharif University of Technology
      \AND
      \name Mohamadreza Fereydooni \email mrezafereydooni@gmail.com \\
      \addr Department of Computer Engineering\\
      Sharif University of Technology
      \AND
      \name Mahdi Ghaznavai \email mahdi.ghaznavi@ce.sharif.edu \\
      \addr Department of Computer Engineering\\
      Sharif University of Technology
      \AND
      \name Mahdieh Soleymani Baghshah \email soleymani@sharif.edu \\
      \addr Department of Computer Engineering\\
      Sharif University of Technology
      }

\def\month{MM}  
\def\year{YYYY} 
\def\openreview{\url{https://openreview.net/forum?id=XXXX}} 

\begin{document}

\maketitle

\begin{abstract}
Recent progress has pushed AI frontiers from pattern-recognition tasks toward problems that require step-by-step, System-2-style reasoning, especially with large language models.
Yet, unlike learning, where generalization and out‑of‑distribution (OOD) evaluation concepts are well formalized, there is no clear, consistent definition or metric for “reasoning ability.”
We propose Complexity Out‑of‑Distribution (Complexity OoD) generalization as a framework and problem setting to define and measure reasoning.
A model exhibits Complexity OoD generalization when it maintains performance on test instances whose minimal required solution complexity, either representational (richer solution structure) or computational (more reasoning steps/program length), exceeds that of all training examples.
We formalize complexity via solution description Kolmogorov complexity and operational proxies (e.g., object/relation counts; reasoning‑step counts), clarifying how Complexity OoD differs from length and compositional OOD.
This lens unifies learning and reasoning: many cases solvable with System‑1‑like processing at low complexity become System‑2‑like under complexity pressure, while System‑2 can be viewed as generalization over solution structures.
We translate this perspective into practice with recommendations for operationalizing Complexity OoD across the stack:
incorporating complexity into benchmark and evaluation metric design
rethinking supervision to target solution traces (from final outcomes to process‑level feedback and RL/search),
seeking and designing inductive biases for Complexity‑OoD generalization,  
addressing learning‑to‑reason spillovers such as spurious shortcuts, semantic robustness, catastrophic forgetting, and step‑wise calibration.
Because Complexity OoD cannot be solved by scaling data alone, progress toward robust reasoning will require architectures and training regimes that explicitly model and allocate computation with respect to complexity.
\end{abstract}

\section{Introduction}

What do the concepts of intelligence, thinking, and specifically reasoning mean, and by what criteria can we confidently assert that one agent possesses superior reasoning ability compared to another?
In parallel with these philosophical inquiries, cognitive science introduces the distinction between System-1 and System-2 thinking \cite{Kahneman11,Stanovich2000Individual}. System-1 processes are rapid, intuitive, and rely heavily on pattern recognition. Many current AI achievements, particularly in areas like computer vision and NLP, demonstrate strong System-1-like capabilities by excelling at pattern recognition tasks \cite{goyal2022inductive,krizhevsky2012imagenet,vaswani2017attention}.
The evaluation of AI models trained for these tasks has traditionally centered on their ability to generalize to unseen data. This is typically measured by performance on a held-out test set drawn from the same underlying distribution as the training data (often termed in-distribution generalization) \cite{goodfellow2016deep,zhang2017understanding}. However, a growing recognition in the field highlights the crucial importance of out-of-distribution (OOD) generalization \cite{Geirhos2020ShortcutLI, recht2019imagenet, Hendrycks2020TheMF, Arjovsky2019InvariantRM, Gulrajani2020InSM}. This more challenging form of generalization assesses a model's robustness and true understanding by evaluating its performance on data that significantly deviates from its training distribution. Consequently, modern benchmarks and evaluation methodologies are increasingly incorporating OOD splits alongside traditional test sets to better gauge a model's true learning capabilities and its ability to handle novel, real-world scenarios \cite{Koh2020WILDSAB, GagnonAudet2022WOODSBF, Sagawa*2020Distributionally, taori2020measuring}.

\begin{figure}[htbp]

  \centering
    \includegraphics[width=0.825\linewidth,keepaspectratio]{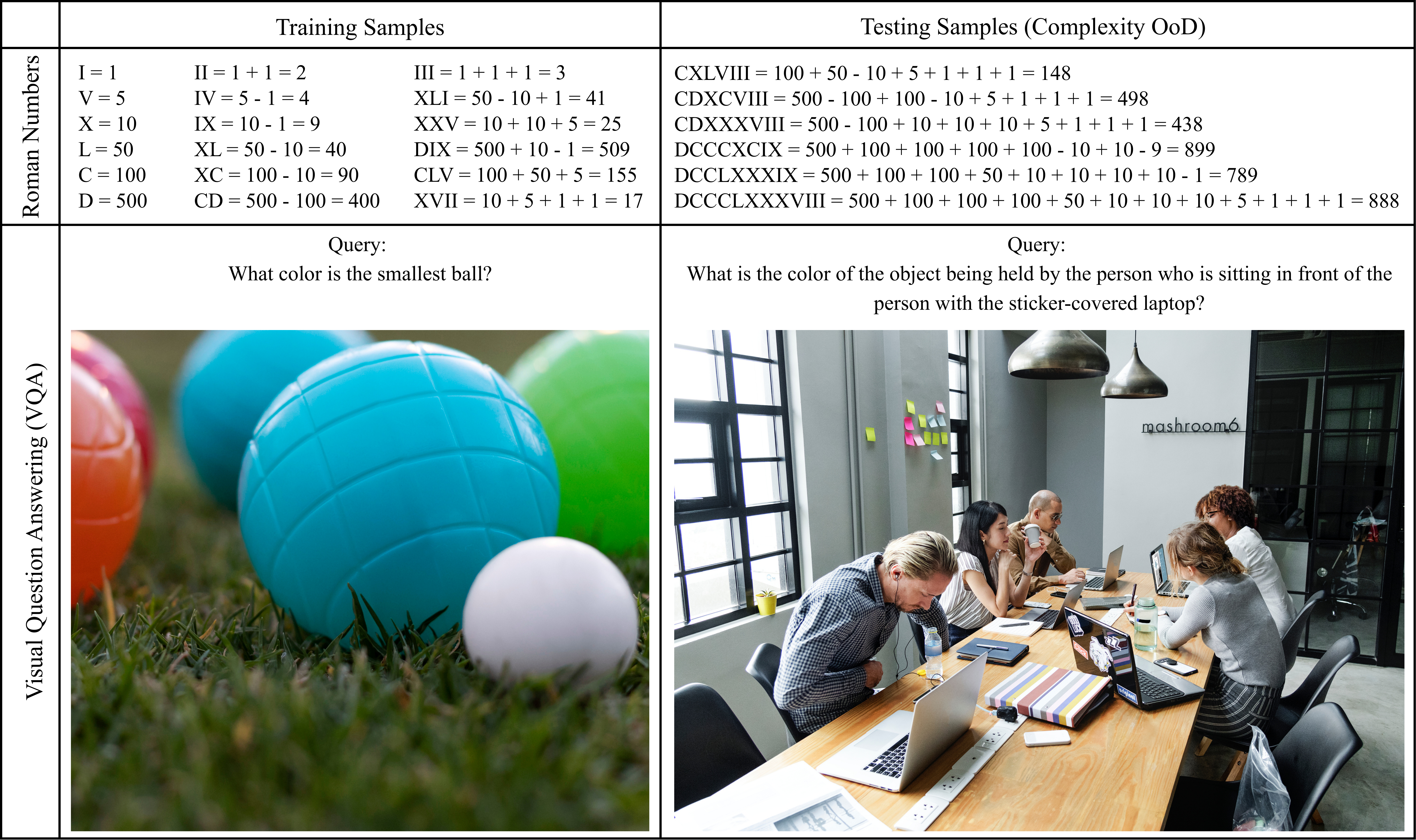}
\caption{Complexity out-of-distribution generalization evaluates whether models trained on problems whose solutions require few, shallow steps generalize to problems whose solutions demand substantially more steps and deeper composition. Two instantiations are shown. Top-Roman numerals: training solutions involve short additive/subtractive decompositions; test solutions require many more operations to expand more complex numerals. Bottom-Visual Question Answering: training features few-hop questions in a simple scene; testing uses relational, multi-hop questions in a busier scene with more entities. The shift is in solution complexity (and, for VQA, scene clutter), not in domain or content.}
    \label{fig:main}


\end{figure}

Alongside these rapid System-1 processes, cognitive science also identifies System-2 thinking. These are characterized by slow, deliberate, and effortful operations, involving analytical thought, complex problem-solving, and crucially, the ability to construct multi-step solutions \cite{Kahneman11,stanovich2011rationality}. These System-2 processes directly correspond to the challenging reasoning tasks that have recently gained significant prominence in the field of artificial intelligence, particularly with the rise of large language models (LLMs) \cite{goyal2022inductive, Li2025FromS1,Wei2022ChainOT}. However, unlike the well-defined metrics of in-distribution and out-of-distribution generalization that evaluate System-1 tasks, there currently lacks a clear and transparent framework for consistently defining and measuring generalization for System-2 reasoning abilities \cite{Mondorf2024BeyondAE,raji2022fallacies}.

In recent years, a proliferation of benchmarks has been introduced, aimed at quantifying the reasoning capabilities of Large Language Models (LLMs) \cite{cobbe2021training, gaoomni, hendrycks2021measuring, srivastava2023beyond,huang2023survey}.
While these models have demonstrated truly impressive performance on many of these tasks, these early benchmarks exhibited several notable limitations. Firstly, their evaluation was often solely predicated on the correctness of the final answer, neglecting the actual reasoning process that led to it \cite{lightman2023lets}. Secondly, and perhaps more critically, while often focusing on narrow domains like mathematics and programming, these benchmarks inadvertently limited their scope, failing to capture the broader, fundamental nature of reasoning itself beyond domain-specific problem-solving \cite{Wei2022ChainOT}. Finally, by failing to account for the underlying distribution of problem instances, they provided an insufficient fine-grained assessment of model performance and inherent limitations, making it difficult to precisely diagnose where and why models struggled \cite{shojaee2025illusionthinkingunderstandingstrengths,taori2023llm}.
More recently, systematic investigations into LLM performance across varying problem difficulties have revealed a critical disconnect: performance on more challenging instances often does not scale proportionally with, or meet the expectations set by, their performance on simpler ones. This observation suggests that models' strong performance on simpler examples might be artificially boosted by data exposure or contamination, blurring the line between genuine reasoning and mere memorization \cite{shojaee2025illusionthinkingunderstandingstrengths, zhou2025gsminfinitellmsbehaveinfinitely, sun2025omegallmsreasonoutside, mirzadeh2025gsmsymbolic, golchin2023trainingsize}. Crucially, this evaluation approach offers minimal insight into the intricate structure or quality of the reasoning traces themselves, making it difficult to truly understand how models arrive at their answers or the robustness of their internal processes \cite{wang2023makes}.
Consequently, the fundamental question of how to reliably discern which model possesses superior reasoning ability, in the absence of a universally accepted definition and robust criteria for genuine reasoning, remains a significant and largely unresolved challenge.

Addressing the persistent challenge of defining and measuring reasoning, this work introduces Complexity Out-of-Distribution (Complexity OoD) generalization as a novel conceptual framework.
Herein, "reasoning ability" is fundamentally reinterpreted as a model's capacity for this specific type of OOD generalization \cite{srivastava2023beyond}. Complexity OOD is formally defined as a scenario where the inherent complexity distribution of test samples significantly surpasses that observed in the training data. Within this framework, 'Complexity' is understood as either the requisite representational capacity or the total number of necessary solution steps for a given problem instance. It is hypothesized that truly superior models are those capable of robustly satisfying this Complexity OOD criterion. The integration of this perspective into evaluation and benchmarking protocols is expected to yield assessments that are markedly more robust against data contamination and offer a more precise, nuanced measure of a model's foundational capabilities \cite{taori2023llm}. Moreover, this framework elucidates a crucial conceptual interplay between conventionally delineated 'learning' (System-1) and 'reasoning' (System-2) tasks \cite{stanovich2011rationality}. It is argued that numerous tasks that typically handled via System‑1 processing, when challenged by instances exhibiting Complexity OOD, inherently evolve into problems demanding a System-2 (reasoning-based) approach for successful resolution. Conversely, by analyzing reasoning through the paradigm of Complexity OOD generalization, it is demonstrated that every System-2 solution can, in turn, be construed as an advanced form of 'Learning' and generalization. This unified perspective aims to bridge the long-standing conceptual divide between learning and reasoning, thereby contributing a more comprehensive framework for comprehending intelligence.

The subsequent sections of this paper elaborate on these contributions. First, a more precise definition of Complexity OOD is provided. This includes differentiating the concept from similar settings, such as compositional OOD, and formally defining it by leveraging principles of Kolmogorov complexity. Next, it is demonstrated how considering Complexity OOD can bridge the concepts of System-2 thinking and learning, revealing that the successful mastery of any System-2 processing often inherently relies on the underlying learning of a System-1-like component. The paper then illustrates that Complexity OOD is not an entirely alien concept within the field; rather, its facets have been observed across various domains, albeit without a unified, overarching perspective. Finally, and most importantly, this work elaborates on the necessary shifts in research methodology and evaluation priorities within the field that arise when assessing models' reasoning abilities through the lens of Complexity OOD.

\section{Complexity Out of Distribution}

\subsection{Motivation}
As previously discussed, cognitive processing can be broadly categorized into two modes: System-1, which is fast and intuitive, and System-2, which is slow and deliberative. This dichotomy is visibly mirrored in the prevailing paradigms of artificial intelligence over the last decade \cite{goyal2022inductive, lowe2024position}.
Tasks addressed within the System-1 framework, such as classifying an image with a fine-tuned ResNet or a piece of text with a BERT model, typically employ an architecture of a fixed computational depth to map an input directly to an output. In this System-1 approach, the primary objective is generalization to unseen samples from the same data distribution or, at best, generalization to a distribution that has undergone a statistical shift \cite{Vapnik1998StatisticalLT, QuioneroCandela2009DatasetSI, hendrycks2018benchmarking, geirhos2020shortcut}.

In contrast, a range of tasks is approached from a System-2 perspective. Examples include solving a mathematical problem with a Large Language Model (LLM)
\cite{wei2022chain}, answering a complex visual query with a Vision-Language Model (VLM)
\cite{Liu2023VisualIT}, or, more abstractly, solving a symbolic regression problem
\cite{biggio2021neural}. In all such cases, "solving" the problem is synonymous with "generating a solution", a coherent sequence of logical sub-steps. The central challenge, therefore, becomes the synthesis of the correct sequence.

The System-2 perspective raises a critical question:
what if a model, ostensibly trained with a System-2 approach, merely "memorizes" the simple and short solution paths present in its training data? Such a model's capacity would be confined to generating solutions of low complexity. Consequently, when faced at test time with an instance requiring a solution path that exceeds this capacity, the model will fail. From the model's perspective, such an instance is out-of-distribution with respect to the complexity of its solution.

We term this scenario Complexity Out-of-Distribution (Complexity OoD) generalization. It dictates that a System-2-based model must be able to generalize over problem instances whose solution complexity is out-of-distribution relative to all training examples. To overcome Complexity OoD, a model must possess a crucial, dynamic capability: the ability to generate a solution of any required complexity on the fly. In other words, during inference, the model must be able to dynamically extend its reasoning process, creating a solution path more complex than any it has seen before.

It is critical to note that the Complexity OoD challenge cannot be resolved merely by scaling training data, as one can always conceive of a test instance with a solution complexity greater than any found in the training set. 
Consequently, achieving Complexity OoD generalization requires the incorporation of appropriate inductive biases.
We posit that if a model can guarantee Complexity OoD generalization for a given task, it can then achieve perfect generalization for any instance of that task.

Finally, the concept of Complexity OoD is not exclusively confined to System-2 approaches. A developer might build a System-1-style model that performs excellently on training examples with limited solution complexity. However, this same model will likely fail when confronted with a test instance that is Complexity OoD, revealing the hidden limitations of its fixed-depth architecture and underscoring the universal importance of this evaluative dimension \cite{Hahn2019TheoreticalLO, Santoro2018MeasuringAR}.

\paragraph{Examples of Complexity OoD.}  
Complexity OoD arises whenever test instances require solutions whose minimal complexity (e.g., number of necessary reasoning steps, proof depth, plan length, or description length) substantially exceeds that of training instances, even when surface statistics remain similar.
As illustrated in ~\autoref{fig:main}, Roman numerals provide a concrete example: The core task involves converting Roman numeral strings to their decimal equivalents (and conversely), adhering to the standard additive–subtractive rules. Elementary numerals (e.g., I, V, X or even II, IV and XX)  can often be processed by System-1 mechanisms, facilitating rapid, intuitive recognition. Conversely, comprehending more intricate numerals like XIX (19), XXIV (24), or LXXXIX (89) mandates the integration of constituent units via a set of compositional rules. This transition moves beyond simple associative recall, requiring System-2 processes to construct systematic solutions by recursively combining previously acquired elements into a coherent representation aligned with the numeral system's structural logic.
Another case arises in arithmetic reasoning. Single-digit multiplications (e.g., $3 \times 4$) can often be recalled directly or processed via System-1 pattern-matching. In contrast, larger multiplications (e.g., $47 \times 89$) require the integration of smaller learned operations into a multi-step algorithm, engaging System-2 processes for systematic computation. \footnote{
For multiplication, LLMs essentially have two distinct approaches: either generating and executing Python code or attempting to perform the calculation internally, without external tools. As highlighted in length generalization studies, a crucial difference emerges: humans, given sufficient attention and working memory, can accurately perform mathematical operations reliably, exhibiting robust length generalization in mathematical reasoning. LLMs, however, do not possess this same guaranteed length generalization, particularly for complex or lengthy mathematical problems when relying solely on internal computation. This is because LLMs, instead of learning the underlying logic of multiplication, a logic inherently generalizable to numbers of any length, primarily learn to mimic the process as observed in their training data. They are, in essence, pattern-matching procedural steps rather than grasping the abstract mathematical principles themselves. This contrasts sharply with human mathematical understanding, which is built upon a foundational grasp of logical structure that ensures generalizability.
}

The applicability of Complexity OoD extends far beyond these foundational symbolic domains, providing a powerful lens for designing and analyzing benchmarks across diverse high‑difficulty tasks. In the realm of visual reasoning and Visual Question Answering (VQA), for example, Complexity OoD can manifest through either increased visual richness or heightened logical demands in the query. A test scene might be significantly more cluttered with objects, attributes, and relations than any training example. Alternatively, the question itself could demand more reasoning hops.
For instance, consider \autoref{fig:main}, a model trained on single-hop questions like “What color is the smallest ball?” could be challenged with a multi-hop query such as “What is the color of the object being held by the person who is sitting in front of the person with the sticker-covered laptop?” Answering this requires a multi-step inferential chain: identify the person with the sticker-covered laptop, determine who is sitting in front of them, detect the object that person is holding, and then report the color of that object.
This principle is equally critical in robotics and long‑horizon planning. A robot might be trained on tasks requiring short action sequences (e.g., “pick up the blue block and place it on the red block”). A Complexity OoD test would demand a significantly longer and more intricate plan, such as “build a four‑block pyramid, which first requires clearing the table by moving all non‑block items into the designated box.” This requires not just more steps, but also managing sub‑goals and interdependent constraints that were absent in the training data. Similarly, in fields like automated theorem proving, a model trained to prove lemmas requiring proofs of a certain depth (e.g., 5–10 inference steps) would face a Complexity OoD challenge when asked to prove a theorem whose shortest proof is an order of magnitude longer. The model must demonstrate an ability to chain inference rules for a duration far exceeding its training experience. The same logic applies to code generation, where a test problem might require programs with greater structural depth, such as more nested functions, intricate recursive patterns, or control flow with deeper nesting, richer branching, and longer dependency chains than any example in the training set. In algorithmic reasoning, this could involve a path-finding model trained on graphs of a certain size being tested on a graph of a similar size but with a significantly larger diameter, forcing the execution of a much longer reasoning sequence. Finally, the concept is highly relevant to narrative and document comprehension. A model may excel at answering questions about short stories where the causal chain is direct and localized. The true test of its reasoning ability, its Complexity OoD performance, comes from processing a long novel and answering a question about a character’s motivation that requires synthesizing subtle clues and events scattered across multiple chapters.

In all these cases, the underlying challenge is the same: the model must dynamically construct a solution or reasoning trace that is structurally more complex than any it has been trained on, moving beyond pattern matching to genuine, scalable procedural understanding.
Across these domains, the common failure mode is not exposure to unfamiliar tokens or images per se, but the need to execute solutions whose minimal complexity exceeds the training support. Conversely, models equipped with inductive biases for adaptive, iterative computation and external memory/tools tend to generalize more gracefully along this axis \cite{Graves2014NTM, Dehghani2019UniversalTransformer, Velickovic2021NeuralAlgorithmicReasoning, Gao2023PAL, Schick2023Toolformer}.

\paragraph{Distinguishing Complexity OoD from Compositionality}  
In the literature on compositional generalization, two primary out-of-distribution scenarios are commonly discussed: \emph{systematicity} and \emph{productivity} \citep{hupkes2020compositionality}. The performance of models has frequently been evaluated under these conditions \citep{lake2018generalization,hupkes2020compositionality,loula2018rearranging}. \emph{Systematicity} refers to the ability to generalize to novel combinations of known components, even when such specific combinations were absent during training \citep{hupkes2020compositionality}. \emph{Productivity}, by contrast, refers to the ability to generalize to sequences of greater length than those encountered during training \citep{hupkes2020compositionality}. Although \emph{complexity OoD} bears some conceptual similarity to compositional OoD, it represents a fundamentally different perspective. The distinction between complexity OoD and systematicity lies in the scope of complexity within the compositions. In systematicity, the challenge is bounded: models must recombine a limited number of familiar primitives. By contrast, complexity OoD imposes no such bound; it emphasizes the need to handle solution paths whose complexity may grow arbitrarily, often requiring deeper reasoning chains characteristic of System-2 processes. The difference between complexity OoD and productivity (often referred to as length OoD) is equally crucial. Length generalization focuses on the size of the input or output sequence, without necessarily implying an increase in reasoning demands. Complexity OoD, however, is defined by the growth of the solution path itself, that is, the number of reasoning or computational steps required to connect input to output. Importantly, these two notions can diverge: a long input may be solvable via a trivial, System-1 style operation, while a short input may require intricate, multi-step System-2 reasoning. For example, a long sequence of repeated symbols (e.g., ``aaaaa...’’) might pose a challenge for productivity but is trivial in terms of reasoning complexity, whereas a short logical puzzle can exemplify complexity OoD by demanding deep multi-step inference despite its brevity.

\paragraph{Representational and Computational Complexity OoD} 
Complexity can be analyzed along two complementary dimensions: the \emph{representational} and the \emph{computational}.  
\emph{Representational complexity OoD} arises when test samples exhibit richer or more intricate structures than those observed during training. Such samples demand finer-grained descriptions or higher-dimensional representations in order to be accurately reconstructed or discriminated.  
The second dimension, \emph{computational complexity OoD}, concerns cases in which obtaining the correct solution requires additional reasoning steps compared to the training regime. Here, the challenge lies not in representing the input but in extending the chain of computation, moving beyond shallow System-1 pattern recognition toward adaptive, multi-step System-2 processing.  
These two dimensions are deeply intertwined. Representational complexity often induces computational complexity: a richer input representation may necessitate longer reasoning paths, while deeper computation may reveal or require more expressive representational structures. Rather than treating them as isolated phenomena, it is crucial to view representational and computational complexity as two sides of the same coin, each shaping and amplifying the other.  
Consequently, addressing complexity OoD requires integrated solutions. A successful framework must accommodate \emph{unbounded representational depth}, which refers to the ability to flexibly encode increasingly complex inputs, as well as \emph{adaptive computational depth}, which refers to the capacity to dynamically extend the number of reasoning steps as needed. We argue that achieving robust System-2 solutions hinges on jointly solving both challenges, thereby enabling models to generalize across variable levels of complexity in real-world data.

\subsection{Formal Definition of Complexity OoD}



System-2 reasoning can be understood as the capacity to handle unbounded complexity in both representation and computation. In contrast to System-1 processing, which often relies on rapid pattern matching and shallow templates for frequent inputs, System-2 processing explicitly constructs and manipulates intermediate structure by composing task‑relevant primitives. Here, primitives denote functional building blocks at the current abstraction level—such as symbols, operators, predicates, spans, or object slots—which may be realized as distributed, overlapping representations yet are treated as indivisible within the active composition scheme. Conceptually, System-2 competence requires the ability to assemble these primitives into longer computational chains and richer descriptions on demand, that is, to represent increasingly intricate inputs and to allocate progressively deeper computation when required.

Let us assume a vocabulary of primitives $M = {m_1, m_2, …, m_n}$. These primitives may serve as representation primitives (analogous to words) or computational primitives (analogous to basic operators). A System-2 solution can then be described as constructing a correct program over $M$. In the representational setting, the program is a structured description (for example, a sentence); in the computational setting, it is an executable procedure (for example, an equation or algorithm). If we further assume access to an oracle that determines whether a given program achieves the goal, then solving reduces to searching for the shortest valid program within the space of programs over M.
\footnote{In the following sections, because the term “program” may be confusing or misleading, we will instead use the term “solution.” However. By solution we mean a possibly unbounded-length, stepwise program and procedure built from semantic primitives, in System-2 processing to arrive at an answer.}

To formalize these ideas, we draw on Kolmogorov Complexity \citep{kolmogorov1965three,li2008introduction}. Although uncomputable in practice, it provides a rigorous theoretical lens for distinguishing between representational and computational complexity in System-2 reasoning.  

\subsubsection{Representational Complexity OoD}  

Let $x$ be an input sample (e.g., an image, a sentence, or a structured object). Its representational complexity is defined as the Kolmogorov Complexity of $x$, denoted $K(x)$:  

\begin{equation}
K(x) = \min \{ |p| : U(p) = x \},
\end{equation}

\noindent where $U(p)$ is the output of a universal Turing machine $U$ given program $p$, and $|p|$ is the program’s length (e.g., in bits). Intuitively, $K(x)$ measures the shortest description length of $x$. High values of $K(x)$ indicate that $x$ has rich or intricate structure, requiring more expressive representations.  

A \emph{representational Complexity OoD} scenario occurs when a test sample $x_{\text{test}}$ requires a description longer than that of any training instance:  

\begin{equation}
K(x_{\text{test}}) > \max_{x_{\text{train}} \in D_{\text{train}}} K(x_{\text{train}}).
\end{equation}

In this case, the model must cope with a representational demand that exceeds its training distribution.

\subsubsection{Computational Complexity OoD}  

Let $y$ denote the solution corresponding to input $x$. In System-2 processes, mapping $x \to y$ often requires a multi-step reasoning procedure. We capture the complexity of this procedure via conditional Kolmogorov Complexity, $K(y \mid x)$:  

\begin{equation}
K(y \mid x) = \min \{ |q| : U(x,q) = y \},
\end{equation}

\noindent where $q$ is a program that takes $x$ as input (or encodes it internally) and produces $y$ as output, with $|q|$ denoting its length. A high $K(y \mid x)$ implies that solving for $y$ requires a longer or more intricate computation.  

A \emph{computational Complexity OoD} scenario occurs when a test pair $(x_{\text{test}}, y_{\text{test}})$ demands a solution program of strictly greater complexity than any training example:  

\begin{equation}
K(y_{\text{test}} \mid x_{\text{test}}) > \max_{(x_{\text{train}}, y_{\text{train}}) \in D_{\text{train}}} K(y_{\text{train}} \mid x_{\text{train}}).
\end{equation}

\subsection{Proxies of Complexity OoD}

Kolmogorov Complexity provides a rigorous theoretical lens for defining representational and computational complexity, but it is uncomputable in practice \citep{li2008introduction}. To study Complexity OoD empirically, we rely on practical proxies that approximate these abstract notions.  

\paragraph{An Example of Representational Complexity Proxy.}  
A useful proxy for representational complexity in the visual domain is the number of objects, attributes, and relations present in a scene. For instance, an image of a single isolated object (e.g., a red cube on a plain background) has low representational complexity, whereas a crowded scene containing multiple overlapping entities with interacting attributes (e.g., ``three people sitting at a table surrounded by books and food'') has significantly higher representational complexity \citep{johnson2017clevr}. Importantly, this notion differs from length generalization. In natural language processing, productivity or length OoD refers to the growth of sequence length (e.g., more tokens in a sentence). By contrast, in images the input dimensions remain fixed, and complexity increases not by length but by the richness of semantic content within the same spatial grid.

\begin{figure}[htbp]
  \centering
\includegraphics[width=0.83\linewidth,keepaspectratio]{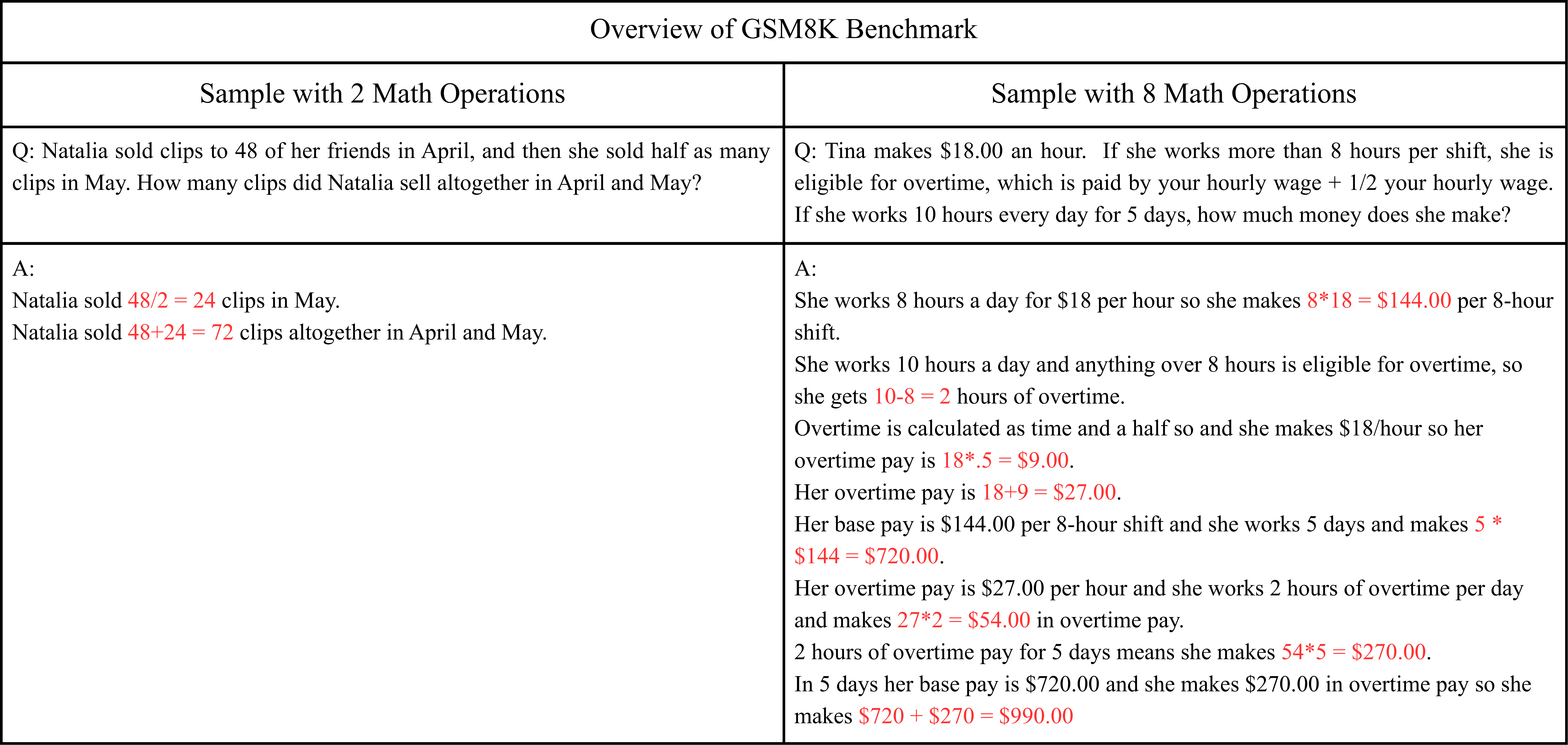}
    \caption{Two examples from the GSM8K dataset in which the number of mathematical operations required to solve the problem can be considered as a proxy for the complexity of the sample problem.}
  \label{fig:gsm-sample}

\end{figure}

\paragraph{An Example  of Computational Complexity Proxy}  
A natural proxy for computational complexity is the length of the reasoning chain required to derive the correct output, as illustrated in Figure~\ref{fig:gsm-sample}. Simple arithmetic such as $2+3$ requires only one step, whereas solving a multi-step algebraic equation or answering a compositional visual reasoning query (e.g., ``Is there a cube to the left of the sphere that is larger than the red object?'') requires a sequence of intermediate inferences \citep{merrill2023lengthgeneralization,wei2022emergent}. Here, the complexity does not arise from longer inputs but from the depth of reasoning steps. This directly reflects the System-2 requirement: solutions must dynamically expand the number of computational steps to accommodate increasingly complex problem instances.  

These proxies make Complexity OoD operational: representational complexity emphasizes the growth of informational richness in inputs, while computational complexity emphasizes the depth of reasoning needed to process them.

\section{The Duality of Learning and Reasoning under Complexity OoD}
The traditional cognitive and AI paradigms often treat learning (System-1) and reasoning (System-2) as distinct, almost modular, faculties. The former is associated with pattern recognition and generalization from data, while the latter is linked to deliberate, multi-step problem-solving. In this section, we argue that the Complexity OoD framework dissolves this rigid dichotomy. It reveals a fundamental duality: many System-1 tasks transform into System-2 challenges when subjected to complexity-out-of-distribution pressures, and conversely, all successful System-2 reasoning can be re-conceptualized as a sophisticated form of learning to generalize over the complexity of solution structures.

\subsection{From Learning to Reasoning: When System-1 Fails, System-2 Emerges}
Consider a canonical System-1 task, such as object recognition. A model trained on images with a few, clearly separated objects learns to map input pixels to labels through what is effectively a high-dimensional pattern-matching function. Its generalization is evaluated on its ability to recognize new objects of the same classes. However, if we evaluate this model on instances with significantly higher representational complexity, for example, scenes with dozens of overlapping interacting objects, the fixed computational path of the model is no longer sufficient. The task is no longer one of simple recognition; it demands a process of segmentation, parsing of inter-object relations, and systematic composition of features, hallmarks of System-2 reasoning. Thus, by pushing a System-1 task into a Complexity OoD regime, we expose its hidden need for a reasoning-based approach

\subsection{From Reasoning to Learning: System-2 as Generalization over Solutions}
Conversely, let us examine a quintessential System-2 task, such as solving a multi-step mathematical problem. As established, a model confronting a computational Complexity OoD instance must generate a solution, with a greater computational depth than any it has observed during training. This perspective, however, naturally raises a critical question: If a model is designed with such a System-2 architecture, how does it "learn" from experience?

The answer lies in reframing the goal. A model that successfully generalizes is not just "thinking" in an abstract sense; it is demonstrating that it has learned a generative procedure for constructing valid solution. The effect of this learning becomes observable along two primary axes:

\begin{itemize}
    \item 
    \textbf{Improved Accuracy}: The most direct form of learning is an increase in the model's ability to generate the correct solution on its first attempt. Through training, its initial output becomes more likely to be valid, reflecting a better-calibrated internal model of the problem space.

    \item 
    \textbf{Improved Efficiency}: A more profound form of learning emerges in settings where a verifier or oracle is available, allowing the model to test its proposed solutions. In such a scenario, learning is not just about being right immediately, but about reaching the correct answer more efficiently, with fewer attempts.
\end{itemize}

From this viewpoint, the solution produced by the model should be seen as a learned heuristic. The model's task is to navigate the vast search space of all possible solutions. A naive or untrained model might engage in a process akin to brute-force search, which is computationally intractable. A trained model, however, learns a heuristic function. This heuristic, itself a product of a System-1-like learning process, guides the construction of the solution by prioritizing more promising paths and pruning the search space.

\begin{itemize}
    \item 
    \textbf{Learning the Primitive Units}: An effective heuristic must operate on a well-defined set of building blocks. As discussed, these atomic units must be sufficient to construct any solution and minimally redundant. Learning them involves an iterative process where units are first tuned on simple tasks and then co-adapted on more complex compositional problems. This provides the System-2 process with a powerful and expressive vocabulary.

    \item 
    \textbf{Learning the Heuristic Function}: With a set of primitive units, the model must learn the heuristic function itself—the policy that dictates how to combine them. This is where the intuition of System-1 plays its most direct role. By being trained on successful and unsuccessful problem-solving traces, the model learns to recognize patterns that predict which sequence of units is most likely to lead to a correct solution. This learned function is what enables the model to bypass exhaustive search and efficiently generate solutions in practice.

\end{itemize}

In this light, the System-2 act of reasoning is powered by a deeply learned System-1-like intuition that guides its deliberate, step-by-step search. Achieving computational Complexity OoD is not an alternative to learning; it is the hallmark of a more profound and robust form of learning, one that masters the underlying structure of solutions, not just the surface statistics of problem-answer pairs. This mastery is what enables the model to generalize its solution-finding process to problems of arbitrary complexity.

\section{Related Works}
We note that from the outset, the field has repeatedly encountered scenarios that implicitly involve shifts in solution complexity, yet these were not framed explicitly as complexity out-of-distribution. In this section, we bring these threads under a single umbrella, our Complexity OoD perspective, and organize the review in three parts. First, we survey work on representational and computational facets of complexity, tracing how variable-length, structured representations and variable-depth computation have been approached (e.g., object-centric and emergent language, adaptive computation, program-synthesize–style methods). Second, we review the recent trajectory in LLMs that integrates reasoning with learning via chain-of-thought prompting, test-time search and deliberation, repeated sampling with self-correction, reward-model supervision, and reinforcement learning. Third, we discuss the emerging trend of complexity-conditioned evaluation that probes long context, compositional structure, exploratory search, and long-horizon execution, advocating complexity-aware reporting instead of single aggregate scores.
\subsection{Variable-length representation.}
\paragraph{Object-Centric Representation Learning:}
Neural networks, particularly those based on conventional convolutional architectures, have demonstrated remarkable success in standard image recognition tasks. Nevertheless, they face pronounced limitations when applied to complex visual scenes comprising multiple objects and intricate inter-object relationships \citep{brady2023provably}. As the complexity of a scene increases, these models often falter due to their inherently fixed-length representational capacity, a limitation known as the superposition catastrophe \citep{von1986thinking, greff2020binding}. This phenomenon refers to the network's inability to disentangle and separately encode multiple entities, resulting in entangled and ambiguous internal representations.

To address these shortcomings, recent research has increasingly focused on object-centric and structured representation approaches, with the Slot Attention mechanism emerging as a prominent example \cite{locatello2020object}. Slot Attention combines low-level perceptual features from convolutional encoders with a fixed set of dynamic “slots” that compete via attention to bind to individual scene elements. Critically, this mechanism supports a flexible number of slots at inference time, allowing it to scale naturally with scene complexity and mitigating the superposition problem by enabling disentangled representations of discrete entities.

Despite these advancements, object-centric representation learning still faces fundamental challenges, particularly in acquiring causal and compositional representations with minimal supervision \cite{didolkar2024zero, mansouriobject, kori2024identifiable, kapl2025object, lekhac2024efficient}. Overcoming these obstacles is essential for the development of more cognitively grounded and System-2-compatible models, highlighting a fertile avenue for future research in artificial intelligence.

\paragraph{Emergent Languages:}
Language, a distinctive hallmark of human cognition, enables intricate communication, complex internal reasoning, and abstract thought. Inspired by these capabilities, researchers have developed the \textit{Emergent Language} paradigm within artificial intelligence \cite{havrylov2017emergence, lazaridou2022multi, lazaridou2018emergence, peters2024emergent}. This field focuses on scenarios wherein multiple artificial agents participate in interactive, game-like tasks that encourage the spontaneous development of structured communication systems. Following the emergence of these novel linguistic forms, researchers analyze their compositional and syntactic properties to assess their functional and cognitive validity \cite{lowe2019pitfalls, chaabouni2020compositionality, carmeli2024evaluating}.

Notably, emergent languages frequently exhibit discrete symbolic units (words) and variable-length message structures, enabling agents to convey information flexibly based on the complexity of their communicative context \cite{ueda2021relationship, lee2024one2many}. This dynamic, context-sensitive flexibility directly aligns with solutions required for System-2 task such as generating detailed, variable-length descriptions based on situational complexity. Furthermore, the discrete, compositional nature of emergent languages closely mirrors the process of generating sophisticated solutions from basic, learnable semantic elements, reinforcing the connection to our conceptualization of processing needed for System-2 problems.

\subsection{Variable-length computation}

\paragraph{Adaptive Computation Time:} One of the fundamental differences between humans and machine learning models is that the human response time to a problem can be a function of the difficulty of that problem, whereas, in machine learning models, the response time solely depends on the model architecture or the size of the input. For example, the longer the input sequence to a recurrent neural network (RNN), the longer it takes for the network to produce the final output. In other words, the human mind can devote more focus and attention to solving a problem with a more challenging input, something that traditional machine learning models are not capable of.
To tackle this issue, Adaptive Computation Time (ACT), a mechanism embedding a halting unit within the RNN architecture, was introduced \cite{graves2016adaptive, chowdhury2024recurrent}. This unit dynamically decides the number of computational steps for each time step by outputting a halting probability, allowing the RNN to either continue processing or move to the next step. This enhancement led to improved performance in tasks like binary vector parity, integer addition, and real number sorting.
The concept of a halting mechanism was extended to the transformer architecture, resulting in the Universal Transformer, which improved performance and accuracy on various algorithmic and language understanding tasks \cite{dehghani2018universal, tan2024sparseuniversal}.

\paragraph{Learning to Program:} Symbolic regression is a problem in machine learning that aims to discover the underlying mathematical expressions or symbolic equations that describe a given dataset. Unlike traditional regression methods that rely on predefined functional forms (based on neural network architecture), symbolic regression attempts to find the symbolic expressions directly from the data.
Symbolic regression has a close relationship with variable-length computation. This relationship arises from the fact that the mathematical expressions discovered by symbolic regression can have varying lengths and complexities, depending on the nature of the underlying relationship in the data \cite{biggio2021neural, kamienny2022end}.
This core idea was later more prominently implemented in the DreamCoder paper
\cite{ellis2021dreamcoder}
. Notably, in DreamCoder, subprograms that frequently co-occurred could be combined and refactored, simplifying the search process across different programs.
Recently, during the 2024 Arc Challenge, a significant number of top-ranked solutions used the Program Generation approach \cite{chollet2024arc, li2024combining, bonnet2024searching,ouellette2024towards, singhal2024conceptsearch}.

\subsection{Some Shines of Integrating System-1 and System-2  via LLMs}
While reasoning problems have traditionally been addressed outside the scope of learning-based methods, recent progress, driven especially by LLMs, has increasingly bridged the gap between the fields of learning and reasoning. More specifically, the generative nature of LLMs enables them to produce variable-length outputs, making them well-suited for tackling reasoning tasks that require flexible, structured solutions. In this section, we introduce recent efforts to solve reasoning problems by leveraging LLMs as a foundational infrastructure.

\paragraph{Chain of Thought (CoT):} 
For reasoning tasks, LLMs can be asked to write the solution step-by-step before providing the final answer \cite{wei2022chain, xia2025beyond}. This can enable the language model to generate longer solutions for more complex problems by generating tokens sequentially.
The CoT idea helped significantly improve the performance of language models on some reasoning tasks. However, since LLMs are still confined to left-to-right decision-making processes (without backtracking) during inference, they can fall short in System-2 tasks that require exploration, strategic lookahead, or where initial decisions play a pivotal role \cite{he2025breaking}. This means that for certain reasoning tasks, LLMs still faced challenges.

\paragraph{LLMs and Search:}
Since the trained LLMs by a System-1 approach can not guarantee to solve all reasoning problems naively by the CoT approach as discussed above, some approaches that need to explore during the test time in order to find the output have been introduced. Ideas such as Tree of Thought (ToT) and Graph of Thought (GoT) allow LLM to branch and generate the solution step-by-step through a search process during the inference time \cite{yao2024tree, besta2023graph, koh2024treesearch, zhang2024restmcts, chen2024alphamath, bi2024forest, yu2024rmcts, wang2025fetch, ding2025dpts}. ToT allows LLMs to perform deliberate decision-making by considering multiple different reasoning paths and self-evaluating choices to decide the next course of action, as well as looking ahead or backtracking when necessary to make global choices. In case of failure, it has the ability to backtrack and construct a new solution. This concept is clearly analogous to the concept of learn-to-search.

\paragraph{LLMs and repeated sampling:} 
LLMs, as probabilistic generative models, offer the capability to generate a diverse range of step-by-step solutions.
LLMs achieve this through \textit{repeated sampling}, a technique that increases the likelihood of generating an optimal response
\cite{Li2022CompetitionlevelCG,Rozire2023CodeLO}. 
Common sampling strategies in LLM inference include top-p (Nucleus Sampling) and top-k sampling, which enable the parallel generation of multiple candidate outputs. By leveraging repeated sampling, LLMs enhance their chances of producing accurate and high-quality responses \cite{brown2024largelanguagemonkeys}, akin to how algorithm designers iteratively refine their solutions to improve computational efficiency.
Self-correction is a test-time computation method that allows LLMs to iteratively revise and refine generated results using external or internal feedback \cite{Shinn2023ReflexionLA, Ye2024PreferenceGuidedRS, Madaan2023SelfRefine}. A critical aspect of this iterative process is the implementation of evaluation and verification strategies, which ensure the effectiveness of repeated sampling and contribute to the overall reliability of the generated outputs. Selecting the most frequent answer as a verification strategy can enhance accuracy, particularly in approaches like self-consistency CoT
\cite{Wang2022SelfConsistencyIC,Li2024MoreAI,Lin2023JustAO}. Moreover, the reward models presented below offer a systematic approach to assessing generated reasoning traces.

\paragraph{LLMs and Reward Models:} Reward models are primarily categorized into two types: Outcome-based Reward Models (ORMs) and Process-based Reward Models (PRMs). ORMs evaluate solutions based solely on the correctness of the final Chain-of-Thought (CoT) output, and thus provide a relatively coarse feedback signal \cite{cobbe2021training, bai2022constitutional}. In contrast, PRMs are trained on finer-grained annotations that assess the validity of each intermediate reasoning step, enabling them to localize errors and provide richer supervisory signals \cite{uesato2022solving, lightman2023lets, wang2023process}. Recent studies show that PRMs significantly outperform ORMs in domains such as mathematics and code generation, as their localized feedback improves both reliability and robustness \cite{lightman2023lets, zhou2024we}. However, PRMs are costly to construct since they require high-quality, step-level annotations, often from domain experts, making scalability a central challenge \cite{wang2023process, shi2024benchmarking}. To alleviate this, automated annotation techniques have been proposed, including Monte Carlo Tree Search (MCTS)-based labeling \cite{wang2024math‌}, synthetic reasoning traces \cite{li2023evaluating}, and weak-to-strong generalization frameworks where smaller, trusted models generate labels for training larger models \cite{burns2023weak}. Importantly, PRMs can also serve as heuristic functions to guide search over candidate reasoning trajectories, closely resembling neural-guided search in program synthesis and theorem proving \cite{polu2022formal, chen2023teaching}. Beyond training, both ORMs and PRMs are increasingly employed at inference time to discriminate between desirable and undesirable outputs, for instance through reranking or rejection sampling across multiple LLM candidates \cite{uesato2022solving, wang2023process, zhou2024we}. This dual utility—providing step-level supervision and enabling inference-time selection—highlights reward models as a key component in advancing the reliability and generalization of reasoning-capable LLMs. On the other hand, reward models can be employed in a Reinforcement Learning (RL) pipeline too as discussed below.

\paragraph{LLMs and RL}:  
A recent approach proposed in several studies \cite{wang2024math‌, setlur2024rewarding, zelikman2022star, huang2023large} is fine-tuning LLMs by an RL paradigm using the CoTs generated by the LLMs themselves and evaluated by verifiers or reward models (mentioned above) which provide supervision feedback. Unlike Supervised Fine-Tuning (SFT), which tends to overfit to training data and struggles with generalization to out-of-distribution scenarios \cite{chu2025sftmemorizesrlgeneralizes, singhal2023towards}, RL methods generalize to unseen situations more effectively by optimizing policies against adaptive reward signals.  
While the community has, to date, often preferred process reward model (PRM)-based verifier methods (especially after the success of the o1 model), several new directions have emerged. For example, DeepSeek R1 \cite{guo2025deepseek} and related work on verifier-free RL \cite{zhou2025verifree} demonstrate that large-scale models can be trained via pure RL with only simple correctness and structural rewards, eliminating explicit verifiers while maintaining competitive performance. Other recent studies propose more efficient reinforcement learning variants, such as contrastive CoT-based reinforced fine-tuning (CARFT) \cite{liu2025carft}, or reinforcement-learning–based knowledge distillation that leverages multi-branch reasoning structures (RLKD) \cite{sun2025rlkd}. Despite their simplicity, these approaches rival verifier-based pipelines like o1 \cite{guo2025deepseek, wang2024math‌, bai2022constitutional}, highlighting the versatility of RL for reasoning.  
From a Complexity OoD perspective, RL approaches are especially significant since they inherently encourage the model to allocate computational resources dynamically in proportion to the complexity of the problem encountered at inference. This enables the emergence of an “aha moment,” where the model recognizes when its initial reasoning path is insufficient for a complex scenario and accordingly invests greater computational effort, revises its reasoning steps, or backtracks to construct a more suitable solution. \cite{zan2025metacognition}.

\subsection{Considering Task Complexity in the Evaluation of LLMs}
After initial successes on short, well‑structured problems, LLMs have very recently been applied to substantially more complex tasks in reasoning, planning, and software engineering, whose defining characteristics include long context, step‑by‑step decision making, and strategic planning. Foundational evidence has already cautioned that Transformers, the backbone of most LLMs, struggle as compositional and structural complexity increases, placing limits on scale‑alone solutions to systematic generalization \cite{dziri2023faith}. Complementing this, \cite{zhou2025gsminfty} systematically scales both context length and reasoning difficulty and observes reliability drop‑offs under increasing length and complexity, motivating compute‑aware protocols and complexity‑conditioned reporting. Complexity‑binned analyses in The \cite{shojaee2025illusionthinkingunderstandingstrengths} further show that models often collapse at higher problem complexity, a pattern consistent with contamination‑inflated performance on easier instances and underscoring the need to evaluate along the complexity axis rather than with a single average. \cite{sinha2025illusiondiminishingreturnsmeasuring} disentangles planning from execution and demonstrates that even when the correct algorithmic plan is provided, models frequently fail over long execution horizons due to brittle state tracking and procedural fidelity, calling for metrics that couple horizon length with step‑level correctness.
In parallel, there is a clear trend toward benchmarks that explicitly probe these complex settings, including \cite{sun2025omegallmsreasonoutside} for exploratory, compositional, and transformative mathematical generalization, \cite{sun2025l0reasoningbenchevaluatingprocedural} for procedural correctness via simple program execution, and \cite{qiu2025locobenchbenchmarklongcontextlarge} for long‑context, cross‑file software engineering with multi‑stage decision making, collectively reinforcing that evaluation of LLMs must be conditioned on task complexity and long‑horizon execution demands.

\section{Constructing the Foundations of Complexity OoD Generalization}

In the preceding sections, we have introduced Complexity OoD as a novel conceptual framework for understanding and evaluating reasoning. We demonstrated how this lens unifies the seemingly disparate concepts of learning (System-1) and reasoning (System-2), revealing a fundamental duality between them. Furthermore, by surveying various research directions, we have shown how the field is already implicitly grappling with different facets of the Complexity OoD challenge.
Beyond this theoretical formulation, however, we must address the practical implications. Having accepted the importance of the Complexity OoD challenge, what changes must we make to our research trajectories, model development practices, and evaluation methodologies? This section proposes several concrete, actionable shifts in perspective and priority that can guide the field toward building more robust and generalizable reasoning agents.

\subsection{Rethinking Evaluation: Tasks and Benchmarks}
System-1 neural network architectures have existed for years, but the rapid evolution of deep learning began with the introduction of the ImageNet dataset in 2012
\cite{Russakovsky2014ImageNetLS}
.
The event referred to as the ImageNet moment made the ImageNet dataset gain significant attention as the first large-scale dataset for benchmarking deep-learning vision networks.
We believe that to ignite the progress of System-2, there must be a spark in creating tasks and benchmarks specifically tailored for it. In other words, System-2 needs its own version of the ImageNet moment.
One example of such a benchmark is the ARC (Abstraction and Reasoning Corpus) Challenge proposed by François Chollet, which consists of tasks designed to evaluate more advanced reasoning capabilities beyond pattern recognition \cite{chollet2019measure}.
Of course, defining a foundational task with maximum inclusivity for System-2 is a non-trivial and complex matter, requiring extensive investigation.
Nevertheless, alongside this main path, a parallel path can be pursued where tasks and benchmarks of System-1 are addressed using an approach inspired by System-2.
For example, consider image classifiers that, upon receiving an image, attempt to generate the output over a variable number of steps depending on the complexity of the image.

\begin{figure}[htbp]
  \centering
  \begin{subfigure}[b]{0.45\linewidth}
    \includegraphics[width=\linewidth]{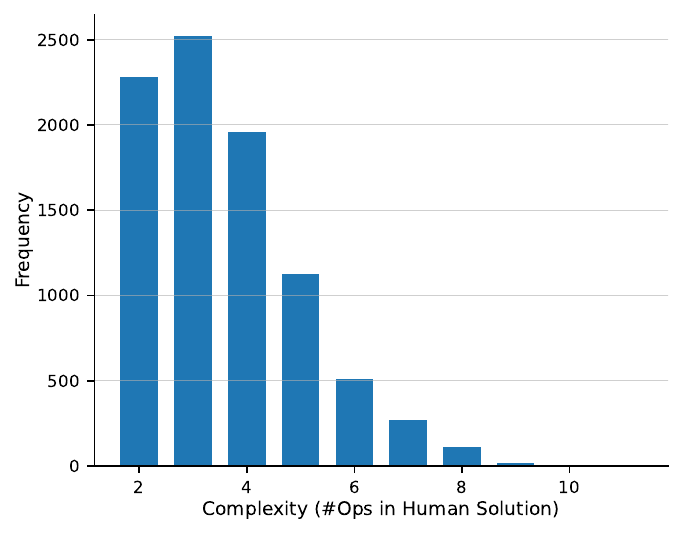}
    \caption{}
    \label{fig:gsm8k-frequency}
  \end{subfigure}
  \begin{subfigure}[b]{0.45\linewidth}
    \includegraphics[width=\linewidth]{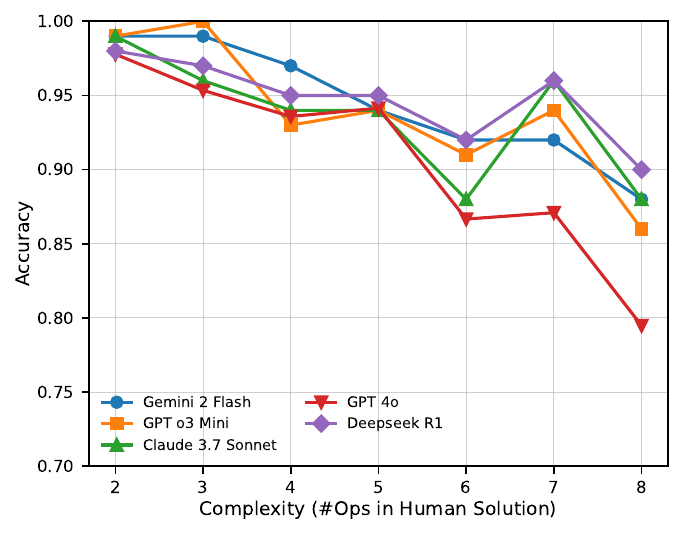}
    \caption{}
    \label{fig:gsm8k-accuracy}
  \end{subfigure}

  \caption{
    (a) The frequency distribution of problem complexity in the GSM8K dataset, measured by the number of arithmetic steps in reference solutions. Most problems are simple, leading to a strong imbalance in the dataset. (b) Model accuracy on GSM8K across different complexity levels, illustrating that as problem complexity increases, model performance drops (especially for non-reasoning models) highlighting the key challenge of complexity out-of-distribution (OoD) generalization. This analysis reveals limitations obscured by average-case metrics and motivates the need for complexity-aware evaluation in benchmarking reasoning ability.
  }
  \label{fig:gsm8k}
\end{figure}

\begin{figure}[h]
  \centering
  \begin{subfigure}[b]{0.45\linewidth}
    \includegraphics[width=\linewidth]{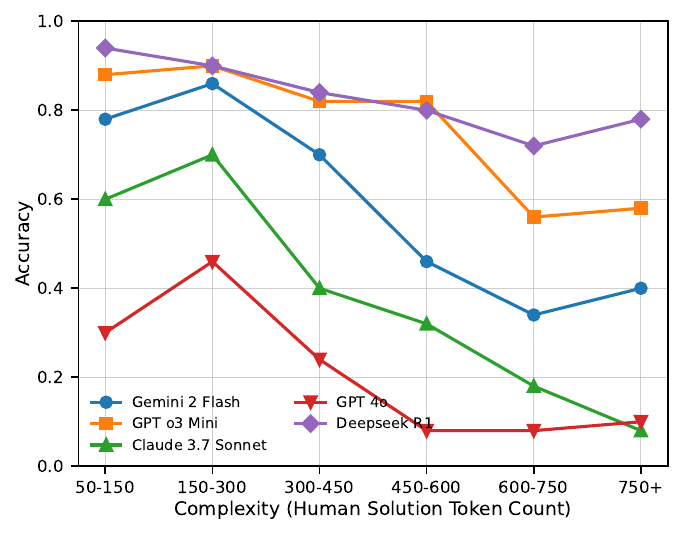}
    \caption{}
    \label{fig:aime}
  \end{subfigure}
  \begin{subfigure}[b]{0.45\linewidth}
    \includegraphics[width=\linewidth]{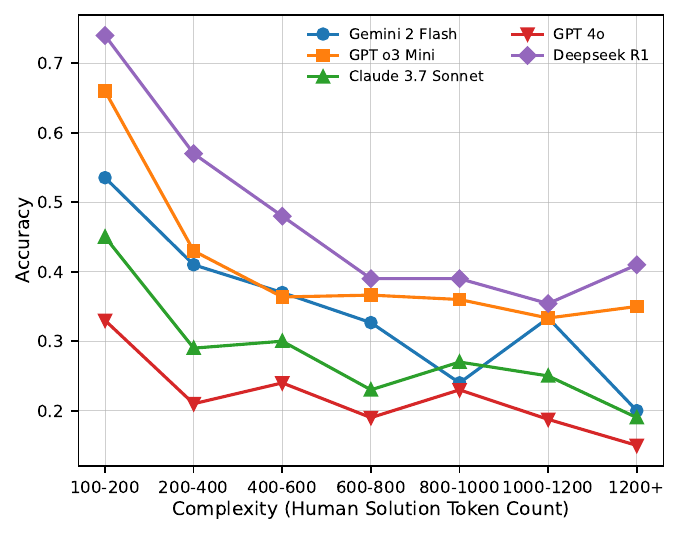}
    \caption{}
    \label{fig:omnimath}
  \end{subfigure}
    \caption{
  (a) Accuracy of Language Models on AIME by Human Solution Token Length.
Complexity is estimated by the number of tokens in provided human-written solutions for each problem, used here as a proxy for the length and intricacy of multi-step reasoning required. Accuracy declines notably with increasing solution length for all models; however, models designed for advanced reasoning (DeepSeek-R1, GPT-o3-mini) maintain higher accuracy and exhibit a gentler degradation compared to more general-purpose models.
(b) Accuracy of Language Models on Omni-MATH by Human Solution Token Length.
.Here too, as the solution complexity (token count) rises, model accuracy drops, especially for models not explicitly optimized for complex reasoning. The pattern reinforces the importance of evaluating models on complexity out-of-distribution (OoD) instances.
  }
  \label{fig:benchmarks}
\end{figure}

Consider the GSM8K dataset, a widely used benchmark consisting of elementary-level math word problems designed to test arithmetic reasoning. While models often achieve high average accuracy on this dataset, it is generally treated as a "solved" benchmark. However, when we analyze models' accuracy based on the inherent complexity of the samples, a different picture emerges.
We define the complexity of each sample as the number of arithmetic operations in its corresponding human-written solution. As shown in Figure~\ref{fig:gsm8k-frequency}, the complexity distribution is highly imbalanced—approximately exponential—where simpler problems are much more frequent than complex ones. Consequently, the standard evaluation metric (average accuracy) is heavily biased toward these simpler cases, potentially providing an overly optimistic view of model capabilities.
As shown in Figure~\ref{fig:gsm8k-accuracy}, when we break down performance by complexity levels, a consistent trend appears: as complexity increases, accuracy decreases. Even though all problems are elementary in nature, LLMs exhibit drops in accuracy for higher-complexity samples. This effect reveals that performance metrics which aggregate over all examples obscure the true generalization capabilities of the model.
Furthermore, the rate at which accuracy deteriorates varies across models. For instance, reasoning-oriented models such as DeepSeek-R1 and GPT-o3-mini show a gentler degradation curve, while non-reasoning models like GPT-4o, break more sharply as complexity rises. This widening performance gap at higher complexity levels reveals an important insight: evaluating models by their failure rate under increasing complexity provides a clearer and more nuanced view of complexity generalization.
Thus, incorporating complexity-aware evaluation into benchmarks like GSM8K highlights the importance of Complexity OoD. It enables us to distinguish between models that merely memorize common patterns and those that demonstrate robust, systematic generalization under increasing reasoning demands.

Similar complexity-aware analysis can be extended to other reasoning benchmarks, such as AIME and Omni-MATH. The AIME dataset consists of problems from the American Invitational Mathematics Examination, featuring high-school level olympiad-style questions that require multi-step reasoning and are more challenging than those in GSM8K. The Omni-MATH dataset goes further, aggregating a wide range of advanced mathematical problems from various national and international mathematics olympiads. These questions are often regarded as some of the most difficult reasoning problems available for benchmarking.
Unlike GSM8K, where complexity can be clearly defined by counting the number of arithmetic operations in a human-written solution, measuring complexity in datasets like AIME and Omni-MATH is more challenging. These datasets often lack standardized, fine-grained, step-by-step human annotations. To approximate reasoning complexity in these cases, we use the number of tokens in the human-written solution as a proxy, interpreting longer solutions as indicative of more elaborate reasoning procedures.
After computing the token lengths for each solution, we group the test samples into buckets of fixed size (e.g., 150 or 200 samples per bin) to equalize comparison and compute accuracy for each complexity bin. As shown in Figure~\ref{fig:benchmarks}, the pattern observed in GSM8K becomes even more pronounced: model accuracy degrades more steeply as problem complexity increases. Importantly, reasoning-oriented models such as DeepSeek-R1 and GPT-o3-mini consistently outperform others across all complexity levels and exhibit a more gradual decline in performance. This aligns with our broader finding that RL-trained models generalize better to complexity OoD cases, supporting the hypothesis that reinforcement learning enhances a model’s ability to dynamically allocate computation and adapt to harder reasoning tasks.
These observations underscore that evaluating model robustness across complexity bins is essential for understanding generalization. Metrics that ignore this aspect fail to reflect true System-2 capabilities. In this light, benchmarks like AIME and Omni-MATH offer critical testbeds for studying models’ behavior in the presence of symbolic complexity, pushing us closer toward evaluating—and achieving—System-2-level generalization.
\subsection{Rethinking Training: Supervision Paradigms}

The challenge of training a System-2 model is fundamentally the challenge of supervising the synthesis of its solution paths. Drawing parallels with System-1, which encompasses supervised, unsupervised, and self-supervised methods, we can categorize the supervisory landscape for reasoning based on the nature and granularity of the available feedback. This distinction is crucial, as the chosen paradigm dictates the scalability of learning and the types of reasoning skills a model can acquire.

\paragraph{Strong Supervision: Learning from Exemplary Solution Traces.}
This is the most direct form of supervision, where the training data consists of triplets: \texttt{(problem, correct solution path, final answer)}. This approach is akin to a student being shown an explicit, step-by-step worked example. It offers a powerful and precise learning signal, making it highly effective for training the solution generation component. However, its primary limitation is the scarcity and high cost of data. Creating high-quality, step-by-step reasoning traces requires significant human expertise and effort, making this paradigm powerful in theory but difficult to scale in practice.

\paragraph{Weak Supervision: Learning from Final Outcomes.}
A far more common and scalable scenario is one where supervision is only available for the final answer, with training data consisting of pairs: \texttt{(problem, final answer)}. In this paradigm, the intermediate reasoning trace is a latent variable that the model must infer. This transforms the learning problem into a difficult credit assignment challenge: if the final answer is wrong, which of the intermediate steps was flawed? This setting is a natural fit for Reinforcement Learning (RL), where the correctness of the final answer serves as a sparse reward signal to guide the exploration of the vast search space of possible solution paths.

\paragraph{Meta-Learning: Learning to Discover Reusable Cognitive Primitives.}
Instead of learning to solve tasks in isolation, we can aim higher: learning to learn how to reason across a diverse range of problems. In this meta-learning paradigm, the model is exposed to a multitude of different tasks. The goal is not just to master each task, but to force the model to discover the shared, underlying atomic units and compositional rules that are useful across all of them. By learning to induce these reusable cognitive primitives, the model acquires a foundational, extensible toolkit for reasoning, enabling it to tackle novel problems more effectively. This aligns with how humans build up a library of problem-solving techniques.

\paragraph{Self-Supervised Learning: Creating Supervision from Unlabeled Data.} Drawing inspiration from the success of self-supervision in System-1, we can devise analogous objectives for learning in the domain of solutions. Given a large corpus of unlabeled solutions or reasoning traces (e.g., from open-source code repositories or scientific papers), we can train the heuristic function for the solution generator. For example, a "masked solution modeling" objective could involve masking a sub-routine or a logical step within a solution and training the model to predict the missing part from the surrounding context. A contrastive objective could train the model to recognize that two different-looking solutions are semantically equivalent (e.g., they implement the same algorithm) or that a slight change to a solution drastically alters its function. Such methods could allow the System-2 machinery to learn the structure and semantics of valid reasoning without requiring paired problem-solution data.

\subsection{Rethinking Methods: Inductive Biases for Complexity Out-of-Distribution}

The remarkable success of deep learning architectures stems from their powerful inductive biases, which align with the inherent structure of data (e.g., translation equivariance in CNNs, sequentiality in RNNs). Just as generalization over any form of out-of-distribution data requires appropriate inductive biases, overcoming the Complexity OoD challenge is fundamentally dependent on them. This necessity is especially critical for Complexity OoD because, as we established earlier, it cannot be resolved merely by scaling training data. For any training set, regardless of the complexity of its instances, one can always construct a test set with problems whose solution complexity exceeds that of the training distribution.

Therefore, the core of any effective inductive bias for Complexity OoD must be to enable unbounded capacity at inference time, for both representation and computation. This means moving beyond architectures with fixed computational graphs and static representational limits. Current models, including the Transformer, possess general-purpose sequence processing capabilities but lack the specific priors needed to efficiently learn and generalize in the structured, combinatorial space of solutions. We identify three crucial areas where new inductive biases are essential to unlock this dynamic, unbounded capacity:

\paragraph{Unbounded Representational Capacity and Solution Structure}: System-1 models operate on fixed-dimensional vectors. In contrast, System-2 reasoning requires representing solutions of variable and potentially unbounded length and complexity. This necessitates a shift from feature spaces to solution spaces. A powerful inductive bias is one that favors modular and compositional representations of solutions. Instead of treating a solution as a flat sequence of tokens, architectures should be biased towards representing them as structured objects like Abstract Syntax Trees (ASTs) or computational graphs. This structural bias would allow the model to learn and reuse sub-routines (functions or modules), a cornerstone of efficient solution synthesis and a key mechanism for generalizing to more complex problems by composing known building blocks in novel ways \cite{ellis2021dreamcoder}.

\paragraph{Adaptive Computational Depth}: A defining feature of human reasoning is the ability to allocate more computational effort to harder problems. Most neural networks, however, have a computational depth fixed by their architecture. To overcome computational Complexity OoD, models must possess an inductive bias for adaptive computation. Mechanisms like the Halting Unit in Adaptive Computation Time (ACT) \cite{graves2016adaptive} or the recurrent nature of Universal Transformers \cite{dehghani2018universal} are early examples. Future research should explore more potent biases for learning recursive and iterative procedures. An architecture with a native bias for recursion could learn the general algorithm for a task (e.g., factorial or graph traversal) from a few examples, enabling it to execute the algorithm for any required depth at inference time, far beyond what was seen during training.

\paragraph{External Memory, Statefulness, and Execution Fidelity}: Complex, multi-step reasoning often requires not only constructing a correct algorithm but also faithfully executing it by meticulously tracking intermediate results and state changes. The human brain relies on working memory for this. Recent studies provide compelling evidence that a core failure mode of LLMs in reasoning tasks is not just an inability to devise a correct algorithm, but an inability to execute one \cite{shojaee2025illusionthinkingunderstandingstrengths, sinha2025illusiondiminishingreturnsmeasuring}. These works demonstrate that even when provided with an explicit, correct algorithm, LLMs often fail by "forgetting" the current state of the problem and proposing invalid actions.\footnote{For instance, when tasked with solving the Tower of Hanoi puzzle, models may correctly follow the rules for several steps but then suggest a move that is physically impossible given the current configuration of disks, indicating a loss of state representation \cite{shojaee2025illusionthinkingunderstandingstrengths}.}

This highlights a critical bottleneck, the transient activations of a Transformer are insufficient to serve as a reliable working memory for complex, stateful procedures. Therefore, an inductive bias for interacting with an external memory structure is not just beneficial, but essential. Architectures like the Neural Turing Machine \cite{Graves2014NTM} provided an early proof-of-concept. By incorporating a bias for reading from and writing to a persistent, stateful memory, a model is no longer required to encode the entire history of its computation within its internal activations. This allows it to offload intermediate products of its reasoning process, freeing up internal resources. More importantly, it enables execution fidelity. Augmenting LLMs with external tools—such as a dedicated memory to store state variables or verifiers that check the validity of each action before execution—can directly address this failure mode. Such a mechanism would empower models to construct and faithfully execute longer and more intricate solution, a prerequisite for overcoming computational Complexity OoD.

\subsection{Rethinking Problems: Revisiting Learning Challenges in the System-2 Domain}

The shift to a System-2 paradigm does not erase the fundamental challenges of machine learning; rather, it recasts them in a new, more abstract domain. As we have argued, System-2 is not a faculty divorced from System-1. Instead, the process of reasoning fundamentally relies on learning: a System-1-like mechanism is learned to act as a heuristic, guiding the search for and construction of a valid solution path.

This deep entanglement means that the System-2 reasoning process inevitably inherits the well-documented "pests" and pathologies of its underlying learning component. Consequently, achieving reliable and trustworthy reasoning is contingent upon our ability to identify, redefine, and address these foundational learning challenges as they manifest in the solution synthesis domain. A robust System-2 agent must be robust not just in its final output, but throughout its entire generation process. In what follows, we illustrate how several canonical learning challenges re-emerge in this new context:

\paragraph{Spurious Correlation and Shortcut Learning in Solution Synthesis}: Shortcut learning in System-2 can be more insidious than in System-1. A model might learn a spurious correlation not between input pixels and a class label, but between superficial textual cues in a problem statement and the structure of the solution. For instance, it might learn that the word "more" always implies an addition operation, failing on problems where "more" is used in a comparative but non-additive context. This is not a failure of calculation, but a failure of learning the correct causal mapping from problem semantics to solution logic. Research inspired by Invariant Risk Minimization \cite{Arjovsky2019InvariantRM} must be adapted to the solution generation context, training models on carefully constructed data distributions that break these spurious links and force the model to learn the underlying algorithm.

\paragraph{Semantic Adversarial Robustness}: Adversarial attacks in System-1 involve small, human-imperceptible perturbations to inputs (e.g., pixels). The equivalent in System-2 is a semantic perturbation: a small, meaning-preserving change to the problem statement that causes a catastrophic failure in the generated solution. For example, changing "Alice has 5 apples, Bob has 3" to "Bob has 3 apples, and Alice has 5" should yield the same solution for a query about the total. A brittle model might be highly sensitive to such word-order variations. Future benchmarks must explicitly test for this semantic robustness, evaluating whether models are learning abstract algorithms or just fragile patterns of text-to-solution mapping.

\paragraph{Catastrophic Forgetting of Reasoning Skills}: In continual learning, catastrophic forgetting occurs when a model trained sequentially on new tasks forgets the knowledge acquired for previous ones. In the System-2 domain, this translates to the forgetting of entire reasoning skills. For instance, a model might be fine-tuned to master geometric proofs, only to exhibit a sharp decline in its previously acquired ability to perform algebraic manipulation. This poses a fundamental obstacle to building cumulative, general-purpose reasoners. If every new skill comes at the cost of an old one, the model can never truly expand its cognitive repertoire. Addressing this requires developing methods for continual learning of algorithmic skills, ensuring that new knowledge is integrated without overwriting foundational abilities.

\paragraph{Poor Calibration and Uncertainty in Multi-Step Reasoning}: In System-1, calibration refers to how well a model's predicted confidence matches its actual correctness. A poorly calibrated model is "confidently wrong." In System-2, this problem becomes more nuanced and critical. A model might generate a multi-step solution, but does it know when it is "stuck" or when a specific reasoning step is likely incorrect? A well-calibrated reasoner should be able to express uncertainty not just about the final answer, but about the intermediate steps of its own reasoning trace. For example, it should be able to signal "I am uncertain about this logical deduction." This lack of self-awareness about its own reasoning process prevents the model from efficient backtracking, asking for help, or strategically allocating more search effort to weaker parts of its solution path.

\paragraph{Algorithmic Bias and Unstated Assumptions}: Bias in System-1 models often manifests as prejudiced outcomes based on sensitive attributes. In System-2, bias can embed itself more subtly within the reasoning process itself, as unstated assumptions. For example, when solving a word problem involving a "doctor" and a "nurse," a model might implicitly assume the doctor is male and the nurse is female based on societal biases in its training data. This assumption, if incorrect for the specific problem, can corrupt the entire reasoning chain, leading to a wrong answer. The problem is not just a biased output, but a biased intermediate step in the algorithm. Combating this requires developing techniques to surface and challenge these implicit assumptions during solution generation.

By explicitly addressing these revisited challenges, we can ensure that our pursuit of System-2 reasoning does not simply replicate the brittleness of System-1 models at a higher level of abstraction, but leads to truly robust and generalizable intelligence.

\section{Conclusion}
In this work, we confronted a fundamental, yet often overlooked, challenge in modern artificial intelligence: the absence of a clear and robust framework for defining and measuring genuine reasoning ability in AI models. We argued that existing evaluation paradigms, largely inherited from System-1 pattern recognition tasks, are insufficient for System-2 reasoning. They often rely on average-case performance metrics that are susceptible to data contamination and fail to provide a fine-grained understanding of a model's true capabilities, particularly its breaking points when faced with novel, complex problems.

To address this gap, we introduced Complexity Out-of-Distribution (Complexity OoD) as a new conceptual framework. We proposed a fundamental reinterpretation: that "reasoning ability" is best understood and measured as a model's capacity to generalize to problem instances whose minimal required solution complexity, be it in representation or computation, lies significantly outside the distribution of its training data. This perspective shifts the focus from simply verifying final answers to assessing the underlying generative process of problem-solving.

The primary power of the Complexity OoD framework is its unifying nature. First, it dissolves the rigid dichotomy between System 1 (learning) and System 2 (reasoning), revealing a profound duality: System-1 tasks become System-2 challenges under complexity pressure, and successful System-2 performance can be reconceptualized as a sophisticated form of learning to generalize over the structure of solution paths. Second, it provides a more robust and diagnostic lens for evaluation, allowing us to move beyond superficial accuracy scores and instead measure how gracefully a model’s performance scales with problem complexity.

Finally, our framework illuminates several concrete and critical future research directions necessary for building the next generation of reasoning agents. These include:

\begin{itemize}
    \item 
Rethinking Benchmarks: We must move towards complexity-aware evaluation, designing benchmarks that explicitly test for Complexity OoD and analyzing model performance across stratified levels of difficulty.

\item
Exploring New Supervision Paradigms: Just as System-1 learning evolved from supervised to self-supervised paradigms, we must explore new forms of supervision for System-2. This involves moving beyond simple outcome-based rewards to process-based supervision, and crucially, developing methods for learning to reason in unsupervised or minimally supervised settings where step-by-step guidance is unavailable.

\item
Inventing New Inductive Biases: Achieving Complexity OoD is not a matter of scale but of architecture. The development of novel inductive biases, such as those for adaptive computation, external memory, and abstraction, is paramount for creating models with the capacity for unbounded reasoning.

\item
Revisiting Foundational Challenges: Classic machine learning problems like spurious correlation, catastrophic forgetting, and adversarial robustness do not disappear; they re-emerge in the domain of solution synthesis and must be redefined and tackled to build truly reliable systems.
\end{itemize}

Achieving System 2-level artificial intelligence will not come from simply scaling up existing models. It demands a fundamental shift in how we evaluate, build, and train models, a shift that equips them with the right inductive biases for generalizing across complexity. By adopting this new lens, we can move beyond measuring performance on static benchmarks and begin to cultivate robust, genuine reasoning. This is the path that will lead to a new generation of AI that does not just learn, but truly thinks.

\clearpage
\bibliography{main}

\begin{thebibliography}{133}
\providecommand{\natexlab}[1]{#1}
\providecommand{\url}[1]{\texttt{#1}}
\expandafter\ifx\csname urlstyle\endcsname\relax
  \providecommand{\doi}[1]{doi: #1}\else
  \providecommand{\doi}{doi: \begingroup \urlstyle{rm}\Url}\fi

\bibitem[Arjovsky et~al.(2019)Arjovsky, Bottou, Gulrajani, and Lopez-Paz]{Arjovsky2019InvariantRM}
Mart{\'i}n Arjovsky, L{\'e}on Bottou, Ishaan Gulrajani, and David Lopez-Paz.
\newblock Invariant risk minimization.
\newblock \emph{ArXiv}, abs/1907.02893, 2019.

\bibitem[Bai et~al.(2022)Bai, Kadavath, Kundu, Askell, Kernion, Jones, Chen, Goldie, Mirhoseini, McKinnon, et~al.]{bai2022constitutional}
Yuntao Bai, Saurav Kadavath, Sandipan Kundu, Amanda Askell, Jason~Phang Kernion, Andrew Jones, Anna Chen, Anna Goldie, Azalia Mirhoseini, Cameron McKinnon, et~al.
\newblock Constitutional ai: Harmlessness from ai feedback.
\newblock \emph{arXiv preprint arXiv:2212.08073}, 2022.

\bibitem[bench authors(2023)]{srivastava2023beyond}
BIG bench authors.
\newblock Beyond the imitation game: Quantifying and extrapolating the capabilities of language models.
\newblock \emph{Transactions on Machine Learning Research}, 2023.
\newblock ISSN 2835-8856.
\newblock URL \url{https://openreview.net/forum?id=uyTL5Bvosj}.

\bibitem[Besta et~al.(2023)Besta, Blach, Kubicek, Gerstenberger, Gianinazzi, Gajda, Lehmann, Podstawski, Niewiadomski, Nyczyk, et~al.]{besta2023graph}
Maciej Besta, Nils Blach, Ales Kubicek, Robert Gerstenberger, Lukas Gianinazzi, Joanna Gajda, Tomasz Lehmann, Michal Podstawski, Hubert Niewiadomski, Piotr Nyczyk, et~al.
\newblock Graph of thoughts: Solving elaborate problems with large language models.
\newblock \emph{arXiv preprint arXiv:2308.09687}, 2023.

\bibitem[Bi et~al.(2024)Bi, Han, Liu, Tang, and Wang]{bi2024forest}
Zhenni Bi, Kai Han, Chuanjian Liu, Yehui Tang, and Yunhe Wang.
\newblock Forest-of-thought: Scaling test-time compute for enhancing llm reasoning.
\newblock \emph{arXiv preprint arXiv:2412.09078}, 2024.
\newblock URL \url{https://arxiv.org/abs/2412.09078}.

\bibitem[Biggio et~al.(2021)Biggio, Bendinelli, Neitz, Lucchi, and Parascandolo]{biggio2021neural}
Luca Biggio, Tommaso Bendinelli, Alexander Neitz, Aurelien Lucchi, and Giambattista Parascandolo.
\newblock Neural symbolic regression that scales.
\newblock In \emph{International Conference on Machine Learning}, pp.\  936--945. Pmlr, 2021.

\bibitem[Bonnet \& Macfarlane(2024)Bonnet and Macfarlane]{bonnet2024searching}
Cl{\'e}ment Bonnet and Matthew~V Macfarlane.
\newblock Searching latent program spaces.
\newblock \emph{arXiv preprint arXiv:2411.08706}, 2024.

\bibitem[Brady et~al.(2023)Brady, Zimmermann, Sharma, Sch{\"o}lkopf, Von~K{\"u}gelgen, and Brendel]{brady2023provably}
Jack Brady, Roland~S Zimmermann, Yash Sharma, Bernhard Sch{\"o}lkopf, Julius Von~K{\"u}gelgen, and Wieland Brendel.
\newblock Provably learning object-centric representations.
\newblock In \emph{International Conference on Machine Learning}, pp.\  3038--3062. PMLR, 2023.

\bibitem[Brown et~al.(2024)Brown, Juravsky, Ehrlich, Clark, Le, Ré, and Mirhoseini]{brown2024largelanguagemonkeys}
Bradley Brown, Jordan Juravsky, Ryan Ehrlich, Ronald Clark, Quoc~V. Le, Christopher Ré, and Azalia Mirhoseini.
\newblock Large language monkeys: Scaling inference compute with repeated sampling, 2024.
\newblock URL \url{https://arxiv.org/abs/2407.21787}.

\bibitem[Burns et~al.(2023)Burns, Nisan, Anil, Chen, Chen, Hilton, Kaplan, Keeling, Kim, McCandlish, et~al.]{burns2023weak}
Collin Burns, Noam Nisan, Cem Anil, Edward Chen, Mark Chen, Jacob Hilton, Jared Kaplan, James Keeling, Jinwoo Kim, Sam McCandlish, et~al.
\newblock Weak-to-strong generalization: Eliciting strong capabilities with weak supervision.
\newblock \emph{arXiv preprint arXiv:2312.09390}, 2023.

\bibitem[Carmeli et~al.(2024)Carmeli, Belinkov, and Meir]{carmeli2024evaluating}
Boaz Carmeli, Yonatan Belinkov, and Ron Meir.
\newblock Evaluating compositionality in emergent communication.
\newblock In \emph{Findings of ACL 2024}, pp.\  189–203, 2024.
\newblock \doi{10.18653/v1/2024.findings-acl.189}.
\newblock URL \url{https://aclanthology.org/2024.findings-acl.189.pdf}.

\bibitem[Chaabouni et~al.(2020)Chaabouni, Kharitonov, Bouchacourt, Dupoux, and Baroni]{chaabouni2020compositionality}
Rahma Chaabouni, Eugene Kharitonov, Diane Bouchacourt, Emmanuel Dupoux, and Marco Baroni.
\newblock Compositionality and generalization in emergent languages.
\newblock In \emph{Proceedings of the 58th Annual Meeting of the Association for Computational Linguistics}, pp.\  4427--4442, 2020.

\bibitem[Chen et~al.(2024)Chen, Liao, Li, and Fan]{chen2024alphamath}
Guoxin Chen, Minpeng Liao, Chengxi Li, and Kai Fan.
\newblock Alphamath almost zero: Process supervision without process.
\newblock In \emph{Advances in Neural Information Processing Systems (NeurIPS)}, 2024.
\newblock URL \url{https://arxiv.org/abs/2405.03553}.

\bibitem[Chen et~al.(2023)Chen, Wang, Chi, Zhou, and Le]{chen2023teaching}
Xinyun Chen, Xuezhi Wang, Ed~Chi, Denny Zhou, and Quoc Le.
\newblock Teaching language models to reason with reinforcement learning.
\newblock In \emph{International Conference on Machine Learning (ICML)}, 2023.

\bibitem[Chollet(2019)]{chollet2019measure}
Fran{\c{c}}ois Chollet.
\newblock On the measure of intelligence.
\newblock \emph{arXiv preprint arXiv:1911.01547}, 2019.

\bibitem[Chollet et~al.(2024)Chollet, Knoop, Kamradt, and Landers]{chollet2024arc}
Francois Chollet, Mike Knoop, Gregory Kamradt, and Bryan Landers.
\newblock Arc prize 2024: Technical report.
\newblock \emph{arXiv preprint arXiv:2412.04604}, 2024.

\bibitem[Chowdhury et~al.(2024)]{chowdhury2024recurrent}
J.~R. Chowdhury et~al.
\newblock Investigating recurrent transformers with dynamic halt.
\newblock \emph{arXiv preprint arXiv:2402.00976}, 2024.
\newblock URL \url{https://arxiv.org/abs/2402.00976}.

\bibitem[Chu et~al.(2025)Chu, Zhai, Yang, Tong, Xie, Schuurmans, Le, Levine, and Ma]{chu2025sftmemorizesrlgeneralizes}
Tianzhe Chu, Yuexiang Zhai, Jihan Yang, Shengbang Tong, Saining Xie, Dale Schuurmans, Quoc~V. Le, Sergey Levine, and Yi~Ma.
\newblock Sft memorizes, rl generalizes: A comparative study of foundation model post-training, 2025.
\newblock URL \url{https://arxiv.org/abs/2501.17161}.

\bibitem[Cobbe et~al.(2021)Cobbe, Kosaraju, Bavarian, Chen, Jun, Kaiser, Plappert, Tworek, Hilton, Nakano, et~al.]{cobbe2021training}
Karl Cobbe, Vineet Kosaraju, Mohammad Bavarian, Mark Chen, Heewoo Jun, Lukasz Kaiser, Matthias Plappert, Jerry Tworek, Jacob Hilton, Reiichiro Nakano, et~al.
\newblock Training verifiers to solve math word problems.
\newblock \emph{arXiv preprint arXiv:2110.14168}, 2021.

\bibitem[Dehghani et~al.(2019{\natexlab{a}})Dehghani, Gouws, Vinyals, Uszkoreit, and Kaiser]{dehghani2018universal}
Mostafa Dehghani, Stephan Gouws, Oriol Vinyals, Jakob Uszkoreit, and Lukasz Kaiser.
\newblock Universal transformers.
\newblock In \emph{International Conference on Learning Representations}, 2019{\natexlab{a}}.
\newblock URL \url{https://openreview.net/forum?id=HyzdRiR9Y7}.

\bibitem[Dehghani et~al.(2019{\natexlab{b}})Dehghani, Gouws, Vinyals, Uszkoreit, and Łukasz Kaiser]{Dehghani2019UniversalTransformer}
Mostafa Dehghani, Stephan Gouws, Oriol Vinyals, Jakob Uszkoreit, and Łukasz Kaiser.
\newblock Universal transformers.
\newblock In \emph{International Conference on Learning Representations (ICLR)}, 2019{\natexlab{b}}.

\bibitem[Didolkar et~al.(2024)Didolkar, Zadaianchuk, Goyal, Mozer, Bengio, Martius, and Seitzer]{didolkar2024zero}
Aniket Didolkar, Andrii Zadaianchuk, Anirudh Goyal, Mike Mozer, Yoshua Bengio, Georg Martius, and Maximilian Seitzer.
\newblock Zero-shot object-centric representation learning.
\newblock \emph{arXiv preprint arXiv:2408.09162}, 2024.

\bibitem[Ding et~al.(2025)Ding, Jiang, Liu, Jing, Guo, Wang, Zhang, Wang, Liu, Du, Liu, and Tao]{ding2025dpts}
Yifu Ding, Wentao Jiang, Shunyu Liu, Yongcheng Jing, Jinyang Guo, Yingjie Wang, Jing Zhang, Zengmao Wang, Ziwei Liu, Bo~Du, Xianglong Liu, and Dacheng Tao.
\newblock Dynamic parallel tree search for efficient llm reasoning.
\newblock In \emph{Proceedings of the 63rd Annual Meeting of the Association for Computational Linguistics (ACL)}, pp.\  11233--11252, 2025.
\newblock \doi{10.18653/v1/2025.acl-long.550}.
\newblock URL \url{https://aclanthology.org/2025.acl-long.550.pdf}.

\bibitem[Dziri et~al.(2023)Dziri, Lu, Sclar, Li, Jiang, Lin, Welleck, West, Bhagavatula, Bras, Hwang, Sanyal, Ren, Ettinger, Harchaoui, and Choi]{dziri2023faith}
Nouha Dziri, Ximing Lu, Melanie Sclar, Xiang~Lorraine Li, Liwei Jiang, Bill~Yuchen Lin, Sean Welleck, Peter West, Chandra Bhagavatula, Ronan~Le Bras, Jena~D. Hwang, Soumya Sanyal, Xiang Ren, Allyson Ettinger, Zaid Harchaoui, and Yejin Choi.
\newblock Faith and fate: Limits of transformers on compositionality.
\newblock In \emph{Thirty-seventh Conference on Neural Information Processing Systems}, 2023.

\bibitem[Ellis et~al.(2021)Ellis, Wong, Nye, Sabl{\'e}-Meyer, Morales, Hewitt, Cary, Solar-Lezama, and Tenenbaum]{ellis2021dreamcoder}
Kevin Ellis, Catherine Wong, Maxwell Nye, Mathias Sabl{\'e}-Meyer, Lucas Morales, Luke Hewitt, Luc Cary, Armando Solar-Lezama, and Joshua~B Tenenbaum.
\newblock Dreamcoder: Bootstrapping inductive program synthesis with wake-sleep library learning.
\newblock In \emph{Proceedings of the 42nd acm sigplan international conference on programming language design and implementation}, pp.\  835--850, 2021.

\bibitem[Gagnon-Audet et~al.(2022)Gagnon-Audet, Ahuja, Bayazi, Dumas, and Rish]{GagnonAudet2022WOODSBF}
Jean-Christophe Gagnon-Audet, Kartik Ahuja, Mohammad Javad~Darvishi Bayazi, G.~Dumas, and Irina Rish.
\newblock Woods: Benchmarks for out-of-distribution generalization in time series tasks.
\newblock \emph{ArXiv}, abs/2203.09978, 2022.
\newblock URL \url{https://api.semanticscholar.org/CorpusID:247594236}.

\bibitem[Gao et~al.()Gao, Song, Yang, Cai, Miao, Dong, Li, Ma, Chen, Xu, et~al.]{gaoomni}
Bofei Gao, Feifan Song, Zhe Yang, Zefan Cai, Yibo Miao, Qingxiu Dong, Lei Li, Chenghao Ma, Liang Chen, Runxin Xu, et~al.
\newblock Omni-math: A universal olympiad level mathematic benchmark for large language models.
\newblock In \emph{The Thirteenth International Conference on Learning Representations}.

\bibitem[Gao et~al.(2023)Gao, Madaan, Zhou, Alon, Liu, Yang, Callan, and Neubig]{Gao2023PAL}
Luyu Gao, Aman Madaan, Shuyan Zhou, Uri Alon, Pengfei Liu, Yiming Yang, Jamie Callan, and Graham Neubig.
\newblock Pal: Program-aided language models.
\newblock In \emph{International Conference on Machine Learning (ICML)}, volume 202 of \emph{Proceedings of Machine Learning Research}, pp.\  10764--10799, 2023.
\newblock URL \url{https://proceedings.mlr.press/v202/gao23f.html}.

\bibitem[Geirhos et~al.(2020{\natexlab{a}})Geirhos, Jacobsen, Michaelis, Zemel, Brendel, Bethge, and Wichmann]{geirhos2020shortcut}
Robert Geirhos, J{\"o}rn-Henrik Jacobsen, Claudio Michaelis, Richard Zemel, Wieland Brendel, Matthias Bethge, and Felix~A Wichmann.
\newblock Shortcut learning in deep neural networks.
\newblock \emph{Nature Machine Intelligence}, 2\penalty0 (11):\penalty0 665--673, 2020{\natexlab{a}}.

\bibitem[Geirhos et~al.(2020{\natexlab{b}})Geirhos, Jacobsen, Michaelis, Zemel, Brendel, Bethge, and Wichmann]{Geirhos2020ShortcutLI}
Robert Geirhos, J{\"o}rn-Henrik Jacobsen, Claudio Michaelis, Richard~S. Zemel, Wieland Brendel, Matthias Bethge, and Felix Wichmann.
\newblock Shortcut learning in deep neural networks.
\newblock \emph{Nature Machine Intelligence}, 2:\penalty0 665 -- 673, 2020{\natexlab{b}}.

\bibitem[Golchin et~al.(2023)Golchin, Zhang, and Liang]{golchin2023trainingsize}
Moein Golchin, Tianyi Zhang, and Percy Liang.
\newblock The effect of training data size on reasoning in large language models.
\newblock \emph{arXiv preprint arXiv:2310.05194}, 2023.
\newblock URL \url{https://arxiv.org/abs/2310.05194}.

\bibitem[Goodfellow et~al.(2016)Goodfellow, Bengio, Courville, and Bengio]{goodfellow2016deep}
Ian Goodfellow, Yoshua Bengio, Aaron Courville, and Yoshua Bengio.
\newblock \emph{Deep learning}, volume~1.
\newblock MIT Press, 2016.

\bibitem[Goyal \& Bengio(2022)Goyal and Bengio]{goyal2022inductive}
Anirudh Goyal and Yoshua Bengio.
\newblock Inductive biases for deep learning of higher-level cognition.
\newblock \emph{Proceedings of the Royal Society A}, 478\penalty0 (2266):\penalty0 20210068, 2022.

\bibitem[Graves(2016)]{graves2016adaptive}
Alex Graves.
\newblock Adaptive computation time for recurrent neural networks.
\newblock \emph{ArXiv}, abs/1603.08983, 2016.
\newblock URL \url{https://api.semanticscholar.org/CorpusID:8224916}.

\bibitem[Graves et~al.(2014)Graves, Wayne, and Danihelka]{Graves2014NTM}
Alex Graves, Greg Wayne, and Ivo Danihelka.
\newblock Neural turing machines.
\newblock In \emph{arXiv preprint arXiv:1410.5401}, 2014.

\bibitem[Greff et~al.(2020)Greff, Van~Steenkiste, and Schmidhuber]{greff2020binding}
Klaus Greff, Sjoerd Van~Steenkiste, and J{\"u}rgen Schmidhuber.
\newblock On the binding problem in artificial neural networks.
\newblock \emph{arXiv preprint arXiv:2012.05208}, 2020.

\bibitem[Gulrajani \& Lopez-Paz(2021)Gulrajani and Lopez-Paz]{Gulrajani2020InSM}
Ishaan Gulrajani and David Lopez-Paz.
\newblock In search of lost domain generalization.
\newblock In \emph{International Conference on Learning Representations}, 2021.
\newblock URL \url{https://openreview.net/forum?id=lQdXeXDoWtI}.

\bibitem[Guo et~al.(2025)Guo, Yang, Zhang, Song, Zhang, Xu, Zhu, Ma, Wang, Bi, et~al.]{guo2025deepseek}
Daya Guo, Dejian Yang, Haowei Zhang, Junxiao Song, Ruoyu Zhang, Runxin Xu, Qihao Zhu, Shirong Ma, Peiyi Wang, Xiao Bi, et~al.
\newblock Deepseek-r1: Incentivizing reasoning capability in llms via reinforcement learning.
\newblock \emph{arXiv preprint arXiv:2501.12948}, 2025.

\bibitem[Hahn(2019)]{Hahn2019TheoreticalLO}
Michael Hahn.
\newblock Theoretical limitations of self-attention in neural sequence models.
\newblock \emph{Transactions of the Association for Computational Linguistics}, 8:\penalty0 156--171, 2019.
\newblock URL \url{https://api.semanticscholar.org/CorpusID:189928186}.

\bibitem[Havrylov \& Titov(2017)Havrylov and Titov]{havrylov2017emergence}
Serhii Havrylov and Ivan Titov.
\newblock Emergence of language with multi-agent games: Learning to communicate with sequences of symbols.
\newblock \emph{Advances in neural information processing systems}, 30, 2017.

\bibitem[He et~al.(2025)]{he2025breaking}
T.~He et~al.
\newblock Breaking the reasoning barrier: A survey on llm complex reasoning.
\newblock \emph{Findings of ACL 2025}, 2025.
\newblock \doi{10.18653/v1/2025.findings-acl.386}.
\newblock URL \url{https://aclanthology.org/2025.findings-acl.386.pdf}.

\bibitem[Hendrycks \& Dietterich(2019)Hendrycks and Dietterich]{hendrycks2018benchmarking}
Dan Hendrycks and Thomas Dietterich.
\newblock Benchmarking neural network robustness to common corruptions and perturbations.
\newblock In \emph{International Conference on Learning Representations}, 2019.
\newblock URL \url{https://openreview.net/forum?id=HJz6tiCqYm}.

\bibitem[Hendrycks et~al.(2020)Hendrycks, Basart, Mu, Kadavath, Wang, Dorundo, Desai, Zhu, Parajuli, Guo, Song, Steinhardt, and Gilmer]{Hendrycks2020TheMF}
Dan Hendrycks, Steven Basart, Norman Mu, Saurav Kadavath, Frank Wang, Evan Dorundo, Rahul Desai, Tyler~Lixuan Zhu, Samyak Parajuli, Mike Guo, Dawn~Xiaodong Song, Jacob Steinhardt, and Justin Gilmer.
\newblock The many faces of robustness: A critical analysis of out-of-distribution generalization.
\newblock \emph{2021 IEEE/CVF International Conference on Computer Vision (ICCV)}, pp.\  8320--8329, 2020.

\bibitem[Hendrycks et~al.(2021)Hendrycks, Burns, Kadavath, Arora, Basart, Tang, Song, and Steinhardt]{hendrycks2021measuring}
Dan Hendrycks, Collin Burns, Saurav Kadavath, Akul Arora, Steven Basart, Eric Tang, Dawn Song, and Jacob Steinhardt.
\newblock Measuring mathematical problem solving with the {MATH} dataset.
\newblock In \emph{Thirty-fifth Conference on Neural Information Processing Systems Datasets and Benchmarks Track (Round 2)}, 2021.
\newblock URL \url{https://openreview.net/forum?id=7Bywt2mQsCe}.

\bibitem[Huang \& Chang(2023)Huang and Chang]{huang2023survey}
Jie Huang and Kevin Chen-Chuan Chang.
\newblock A survey on reasoning in large language models: Taxonomy, evaluation, and future directions.
\newblock \emph{arXiv preprint arXiv:2312.11562}, 2023.
\newblock URL \url{https://arxiv.org/abs/2312.11562}.

\bibitem[Huang et~al.(2023)Huang, Le, Zhou, and Abbeel]{huang2023large}
Yuhuai Huang, Quoc~V Le, Denny Zhou, and Pieter Abbeel.
\newblock Large language models are reinforcement learners.
\newblock \emph{arXiv preprint arXiv:2306.15195}, 2023.

\bibitem[Hupkes et~al.(2020)Hupkes, Dankers, Mul, and Bruni]{hupkes2020compositionality}
Dieuwke Hupkes, Verna Dankers, Mathijs Mul, and Elia Bruni.
\newblock Compositionality decomposed: How do neural networks generalise?
\newblock \emph{Journal of Artificial Intelligence Research}, 67:\penalty0 757--795, 2020.

\bibitem[Johnson et~al.(2017)Johnson, Hariharan, van~der Maaten, Fei-Fei, Zitnick, and Girshick]{johnson2017clevr}
Justin Johnson, Bharath Hariharan, Laurens van~der Maaten, Li~Fei-Fei, C.~Lawrence Zitnick, and Ross Girshick.
\newblock Clevr: A diagnostic dataset for compositional language and elementary visual reasoning.
\newblock In \emph{CVPR}, 2017.

\bibitem[Kahneman(2011)]{Kahneman11}
Daniel Kahneman.
\newblock \emph{Thinking, Fast and Slow}.
\newblock Farrar, Straus and Giroux, New York, 2011.
\newblock ISBN 978-0-374-27563-1.

\bibitem[Kamienny et~al.(2022)Kamienny, d'Ascoli, Lample, and Charton]{kamienny2022end}
Pierre-Alexandre Kamienny, St{\'e}phane d'Ascoli, Guillaume Lample, and Fran{\c{c}}ois Charton.
\newblock End-to-end symbolic regression with transformers.
\newblock \emph{Advances in Neural Information Processing Systems}, 35:\penalty0 10269--10281, 2022.

\bibitem[Kapl et~al.(2025)Kapl, Karimi~Mamaghan, Horn, Marr, Bauer, and Dittadi]{kapl2025object}
Ferdinand Kapl, Amir~Mohammad Karimi~Mamaghan, Max Horn, Carsten Marr, Stefan Bauer, and Andrea Dittadi.
\newblock Object-centric representations generalize better compositionally with less compute.
\newblock \emph{arXiv preprint arXiv:2501.04650}, 2025.
\newblock URL \url{https://arxiv.org/abs/2501.04650}.

\bibitem[Koh et~al.(2024)Koh, McAleer, Fried, and Salakhutdinov]{koh2024treesearch}
Jing~Yu Koh, Stephen McAleer, Daniel Fried, and Ruslan Salakhutdinov.
\newblock Tree search for language model agents.
\newblock \emph{arXiv preprint arXiv:2407.01476}, 2024.
\newblock URL \url{https://arxiv.org/abs/2407.01476}.

\bibitem[Koh et~al.(2020)Koh, Sagawa, Marklund, Xie, Zhang, Balsubramani, Hu, Yasunaga, Phillips, Gao, Lee, David, Stavness, Guo, Earnshaw, Haque, Beery, Leskovec, Kundaje, Pierson, Levine, Finn, and Liang]{Koh2020WILDSAB}
Pang~Wei Koh, Shiori Sagawa, Henrik Marklund, Sang~Michael Xie, Marvin Zhang, Akshay Balsubramani, Weihua Hu, Michihiro Yasunaga, Richard~Lanas Phillips, Irena Gao, Tony Lee, Etiene David, Ian Stavness, Wei Guo, Berton~A. Earnshaw, Imran~S. Haque, Sara~Meghan Beery, Jure Leskovec, Anshul Kundaje, Emma Pierson, Sergey Levine, Chelsea Finn, and Percy Liang.
\newblock Wilds: A benchmark of in-the-wild distribution shifts.
\newblock In \emph{International Conference on Machine Learning}, 2020.

\bibitem[Kolmogorov(1965)]{kolmogorov1965three}
Andrei~N Kolmogorov.
\newblock Three approaches to the quantitative definition ofinformation’.
\newblock \emph{Problems of information transmission}, 1\penalty0 (1):\penalty0 1--7, 1965.

\bibitem[Kori et~al.(2024)Kori, Locatello, Santhirasekaram, Toni, Glocker, and De~Sousa~Ribeiro]{kori2024identifiable}
Avinash Kori, Francesco Locatello, Ainkaran Santhirasekaram, Francesca Toni, Ben Glocker, and Fabio De~Sousa~Ribeiro.
\newblock Identifiable object-centric representation learning via probabilistic slot attention.
\newblock In \emph{Advances in Neural Information Processing Systems (NeurIPS)}, 2024.
\newblock URL \url{https://arxiv.org/abs/2405.05678}.

\bibitem[Krizhevsky et~al.(2012)Krizhevsky, Sutskever, and Hinton]{krizhevsky2012imagenet}
Alex Krizhevsky, Ilya Sutskever, and Geoffrey~E. Hinton.
\newblock Imagenet classification with deep convolutional neural networks.
\newblock In \emph{Advances in Neural Information Processing Systems}, volume~25, pp.\  1097--1105. Curran Associates, Inc., 2012.
\newblock \doi{10.1145/3065386}.
\newblock URL \url{https://proceedings.neurips.cc/paper/2012/hash/c399862d3b9d6b76c8436e924a68c45b-Abstract.html}.

\bibitem[Lake \& Baroni(2018)Lake and Baroni]{lake2018generalization}
Brenden Lake and Marco Baroni.
\newblock Generalization without systematicity: On the compositional skills of sequence-to-sequence recurrent networks.
\newblock In \emph{International conference on machine learning}, pp.\  2873--2882. PMLR, 2018.

\bibitem[Lazaridou et~al.(2018)Lazaridou, Hermann, Tuyls, and Clark]{lazaridou2018emergence}
Angeliki Lazaridou, Karl~Moritz Hermann, Karl Tuyls, and Stephen Clark.
\newblock Emergence of linguistic communication from referential games with symbolic and pixel input.
\newblock In \emph{International Conference on Learning Representations}, 2018.

\bibitem[Lazaridou et~al.(2022)Lazaridou, Peysakhovich, and Baroni]{lazaridou2022multi}
Angeliki Lazaridou, Alexander Peysakhovich, and Marco Baroni.
\newblock Multi-agent cooperation and the emergence of (natural) language.
\newblock In \emph{International Conference on Learning Representations}, 2022.

\bibitem[Le~Khac et~al.(2024)Le~Khac, Healy, and Smeaton]{lekhac2024efficient}
Phuc~H. Le~Khac, Graham Healy, and Alan~F. Smeaton.
\newblock Efficient object-centric representation learning with pre-trained geometric prior.
\newblock \emph{arXiv preprint arXiv:2412.12331}, 2024.
\newblock URL \url{https://arxiv.org/abs/2412.12331}.

\bibitem[Lee et~al.(2024)]{lee2024one2many}
Hyeonwoo Lee et~al.
\newblock One-to-many communication and compositionality in emergent communication.
\newblock In \emph{EMNLP 2024}, pp.\  –, 2024.
\newblock \doi{10.18653/v1/2024.emnlp-main.1157}.
\newblock URL \url{https://aclanthology.org/2024.emnlp-main.1157.pdf}.

\bibitem[Li et~al.(2024{\natexlab{a}})Li, Zhang, Yu, Fu, and Ye]{Li2024MoreAI}
Junyou Li, Qin Zhang, Yangbin Yu, Qiang Fu, and Deheng Ye.
\newblock More agents is all you need.
\newblock \emph{ArXiv}, abs/2402.05120, 2024{\natexlab{a}}.

\bibitem[Li et~al.(2008)Li, Vit{\'a}nyi, et~al.]{li2008introduction}
Ming Li, Paul Vit{\'a}nyi, et~al.
\newblock \emph{An introduction to Kolmogorov complexity and its applications}, volume~3.
\newblock Springer, 2008.

\bibitem[Li et~al.(2024{\natexlab{b}})Li, Hu, Larsen, Wu, Alford, Woo, Dunn, Tang, Naim, Nguyen, et~al.]{li2024combining}
Wen-Ding Li, Keya Hu, Carter Larsen, Yuqing Wu, Simon Alford, Caleb Woo, Spencer~M Dunn, Hao Tang, Michelangelo Naim, Dat Nguyen, et~al.
\newblock Combining induction and transduction for abstract reasoning.
\newblock \emph{arXiv preprint arXiv:2411.02272}, 2024{\natexlab{b}}.

\bibitem[Li et~al.(2023)Li, Madaan, Zhou, Gu, Zhang, Yao, Zhong, Yu, and Zhou]{li2023evaluating}
Xinyun Li, Aman Madaan, Shuyan Zhou, Yewen Gu, Hattie~Zhou Zhang, Shunyu Yao, Victor Zhong, Ping Yu, and Denny Zhou.
\newblock Evaluating process supervision for llm reasoning.
\newblock \emph{arXiv preprint arXiv:2310.02338}, 2023.

\bibitem[Li et~al.(2022)Li, Choi, Chung, Kushman, Schrittwieser, Leblond, Tom, Eccles, Keeling, Gimeno, Lago, Hubert, Choy, de, d’Autume, Babuschkin, Chen, Huang, Welbl, Gowal, Alexey, Cherepanov, Molloy, Mankowitz, Robson, Kohli, de, Freitas, Kavukcuoglu, and Vinyals]{Li2022CompetitionlevelCG}
Yujia Li, David Choi, Junyoung Chung, Nate Kushman, Julian Schrittwieser, R{\'e}mi Leblond, Tom, Eccles, James Keeling, Felix Gimeno, Agustin~Dal Lago, Thomas Hubert, Peter Choy, Cyprien de, Masson d’Autume, Igor Babuschkin, Xinyun Chen, Po-Sen Huang, Johannes Welbl, Sven Gowal, Alexey, Cherepanov, James Molloy, Daniel~Jaymin Mankowitz, Esme~Sutherland Robson, Pushmeet Kohli, Nando de, Freitas, Koray Kavukcuoglu, and Oriol Vinyals.
\newblock Competition-level code generation with alphacode.
\newblock \emph{Science}, 378:\penalty0 1092 -- 1097, 2022.

\bibitem[Li et~al.(2025)Li, Zhang, Zhang, Zhang, Liu, Yao, Xu, Zheng, Wang, Chen, Zhang, Yin, Dong, Guo, Song, and Liu]{Li2025FromS1}
Zhong-Zhi Li, Duzhen Zhang, Ming-Liang Zhang, Jiaxin Zhang, Zengyan Liu, Yuxuan Yao, Haotian Xu, Junhao Zheng, Pei-Jie Wang, Xiuyi Chen, Yingying Zhang, Fei Yin, Jiahua Dong, Zhijiang Guo, Le~Song, and Cheng-Lin Liu.
\newblock From system 1 to system 2: A survey of reasoning large language models.
\newblock \emph{ArXiv}, abs/2502.17419, 2025.

\bibitem[Lightman et~al.(2023)Lightman, Nye, Tenenbaum, and Andreas]{lightman2023lets}
Ben Lightman, Maxwell Nye, Joshua~B. Tenenbaum, and Jacob Andreas.
\newblock Let's verify step by step.
\newblock In \emph{Advances in Neural Information Processing Systems (NeurIPS)}, volume~36, 2023.
\newblock URL \url{https://proceedings.neurips.cc/paper_files/paper/2023/hash/1c38192f7b9d7a3da7e3d75a6b7b2c46-Abstract-Conference.html}.

\bibitem[Lin et~al.(2023)Lin, Fu, Liu, Li, Gong, Wan, Zhang, Wang, Zhang, and Gai]{Lin2023JustAO}
Lei Lin, Jiayi Fu, Pengli Liu, Qingyang Li, Yan Gong, Junchen Wan, Fuzheng Zhang, Zhongyuan Wang, Di~Zhang, and Kun Gai.
\newblock Just ask one more time! self-agreement improves reasoning of language models in (almost) all scenarios.
\newblock In \emph{Annual Meeting of the Association for Computational Linguistics}, 2023.

\bibitem[Liu et~al.(2023)Liu, Li, Wu, and Lee]{Liu2023VisualIT}
Haotian Liu, Chunyuan Li, Qingyang Wu, and Yong~Jae Lee.
\newblock Visual instruction tuning.
\newblock \emph{ArXiv}, abs/2304.08485, 2023.
\newblock URL \url{https://api.semanticscholar.org/CorpusID:258179774}.

\bibitem[Liu et~al.(2025)Liu, Zhang, Chen, Li, and Zhao]{liu2025carft}
Yichen Liu, Wei Zhang, Haoran Chen, Ming Li, and Tianqi Zhao.
\newblock Carft: Contrastive learning with annotated chain-of-thought reinforced fine-tuning.
\newblock \emph{arXiv preprint arXiv:2508.15868}, 2025.

\bibitem[Locatello et~al.(2020)Locatello, Weissenborn, Unterthiner, Mahendran, Heigold, Uszkoreit, Dosovitskiy, and Kipf]{locatello2020object}
Francesco Locatello, Dirk Weissenborn, Thomas Unterthiner, Aravindh Mahendran, Georg Heigold, Jakob Uszkoreit, Alexey Dosovitskiy, and Thomas Kipf.
\newblock Object-centric learning with slot attention.
\newblock \emph{Advances in Neural Information Processing Systems}, 33:\penalty0 11525--11538, 2020.

\bibitem[Loula et~al.(2018)Loula, Baroni, and Lake]{loula2018rearranging}
Joao Loula, Marco Baroni, and Brenden~M Lake.
\newblock Rearranging the familiar: Testing compositional generalization in recurrent networks.
\newblock \emph{arXiv preprint arXiv:1807.07545}, 2018.

\bibitem[Lowe et~al.(2019)Lowe, Foerster, Boureau, Pineau, and Dauphin]{lowe2019pitfalls}
Ryan Lowe, Jakob Foerster, Y-Lan Boureau, Joelle Pineau, and Yann Dauphin.
\newblock On the pitfalls of measuring emergent communication.
\newblock In \emph{Proceedings of the 18th International Conference on Autonomous Agents and MultiAgent Systems}, pp.\  693--701, 2019.

\bibitem[Lowe(2024)]{lowe2024position}
Scott~C. Lowe.
\newblock Position: System-2 reasoning capabilities are nigh.
\newblock In \emph{The First Workshop on System-2 Reasoning at Scale, NeurIPS'24}, 2024.
\newblock URL \url{https://openreview.net/forum?id=ZxaipE6f62}.

\bibitem[Madaan et~al.(2023)Madaan, Tandon, Gupta, Hallinan, Gao, Wiegreffe, Alon, Dziri, Prabhumoye, Yang, Gupta, Majumder, Hermann, Welleck, Yazdanbakhsh, and Clark]{Madaan2023SelfRefine}
Aman Madaan, Niket Tandon, Prakhar Gupta, Skyler Hallinan, Luyu Gao, Sarah Wiegreffe, Uri Alon, Nouha Dziri, Shrimai Prabhumoye, Yiming Yang, Shashank Gupta, Bodhisattwa~Prasad Majumder, Katherine Hermann, Sean Welleck, Amir Yazdanbakhsh, and Peter Clark.
\newblock Self-refine: Iterative refinement with self-feedback.
\newblock In \emph{Advances in Neural Information Processing Systems (NeurIPS) 2023 — Poster / OpenReview}, 2023.
\newblock URL \url{https://openreview.net/forum?id=S37hOerQLB}.
\newblock Poster / OpenReview version.

\bibitem[Mansouri et~al.(2024)Mansouri, Hartford, Zhang, and Bengio]{mansouriobject}
Amin Mansouri, Jason Hartford, Yan Zhang, and Yoshua Bengio.
\newblock Object-centric architectures enable efficient causal representation learning.
\newblock In \emph{The Twelfth International Conference on Learning Representations (ICLR 2024)}, 2024.
\newblock URL \url{https://openreview.net/forum?id=r9FsiXZxZt}.
\newblock ICLR 2024 — Poster.

\bibitem[Merrill et~al.(2023)Merrill, Sabharwal, and Schwartz]{merrill2023lengthgeneralization}
William Merrill, Ashish Sabharwal, and Roy Schwartz.
\newblock The formal limitations of transformer language models at length generalization.
\newblock \emph{Transactions of the Association for Computational Linguistics}, 11:\penalty0 1273--1290, 2023.

\bibitem[Mirzadeh et~al.(2025)Mirzadeh, Alizadeh, Shahrokhi, Tuzel, Bengio, and Farajtabar]{mirzadeh2025gsmsymbolic}
Seyed~Iman Mirzadeh, Keivan Alizadeh, Hooman Shahrokhi, Oncel Tuzel, Samy Bengio, and Mehrdad Farajtabar.
\newblock {GSM}-symbolic: Understanding the limitations of mathematical reasoning in large language models.
\newblock In \emph{The Thirteenth International Conference on Learning Representations}, 2025.

\bibitem[Mondorf \& Plank(2024)Mondorf and Plank]{Mondorf2024BeyondAE}
Philipp Mondorf and Barbara Plank.
\newblock Beyond accuracy: Evaluating the reasoning behavior of large language models - a survey.
\newblock \emph{ArXiv}, abs/2404.01869, 2024.

\bibitem[Ouellette(2024)]{ouellette2024towards}
Simon Ouellette.
\newblock Towards efficient neurally-guided program induction for arc-agi.
\newblock \emph{arXiv preprint arXiv:2411.17708}, 2024.

\bibitem[Peters et~al.(2025)Peters, Waubert~de Puiseau, Tercan, Gopikrishnan, Lucas~de Carvalho, Bitter, and Meisen]{peters2024emergent}
Jannik Peters, Constantin Waubert~de Puiseau, Hasan Tercan, Arya Gopikrishnan, Gustavo~Adolpho Lucas~de Carvalho, Christian Bitter, and Tobias Meisen.
\newblock Emergent language: a survey and taxonomy.
\newblock \emph{Autonomous Agents and Multi-Agent Systems}, 39\penalty0 (1), March 2025.
\newblock ISSN 1573-7454.
\newblock \doi{10.1007/s10458-025-09691-y}.
\newblock URL \url{http://dx.doi.org/10.1007/s10458-025-09691-y}.

\bibitem[Polu \& Han(2022)Polu and Han]{polu2022formal}
Stanislas Polu and Jesse~Michael Han.
\newblock Formal mathematics statement curriculum learning.
\newblock In \emph{International Conference on Learning Representations (ICLR)}, 2022.

\bibitem[Qiu et~al.(2025)Qiu, Liu, Liu, Murthy, Zhang, Chen, Wang, Zhu, Yang, Tan, Cen, Qian, Heinecke, Yao, Savarese, Xiong, and Wang]{qiu2025locobenchbenchmarklongcontextlarge}
Jielin Qiu, Zuxin Liu, Zhiwei Liu, Rithesh Murthy, Jianguo Zhang, Haolin Chen, Shiyu Wang, Ming Zhu, Liangwei Yang, Juntao Tan, Zhepeng Cen, Cheng Qian, Shelby Heinecke, Weiran Yao, Silvio Savarese, Caiming Xiong, and Huan Wang.
\newblock Locobench: A benchmark for long-context large language models in complex software engineering, 2025.
\newblock URL \url{https://arxiv.org/abs/2509.09614}.

\bibitem[Quionero-Candela et~al.(2009)Quionero-Candela, Sugiyama, Schwaighofer, and Lawrence]{QuioneroCandela2009DatasetSI}
Joaquin Quionero-Candela, Masashi Sugiyama, Anton Schwaighofer, and Neil~D. Lawrence.
\newblock Dataset shift in machine learning.
\newblock 2009.
\newblock URL \url{https://api.semanticscholar.org/CorpusID:61294087}.

\bibitem[Raji et~al.(2022)Raji, Taori, Schmidt, and Recht]{raji2022fallacies}
Inioluwa~Deborah Raji, Rohan Taori, Ludwig Schmidt, and Benjamin Recht.
\newblock Fallacies of distributional evaluation: Towards a unified framework for reasoning generalization.
\newblock In \emph{NeurIPS 2022 Workshop on Distribution Shifts}, 2022.
\newblock URL \url{https://arxiv.org/abs/2210.01963}.

\bibitem[Recht et~al.(2019)Recht, Roelofs, Schmidt, and Shankar]{recht2019imagenet}
Benjamin Recht, Rebecca Roelofs, Ludwig Schmidt, and Vaishaal Shankar.
\newblock Do imagenet classifiers generalize to imagenet?
\newblock In \emph{International conference on machine learning}, pp.\  5389--5400. PMLR, 2019.

\bibitem[Rozi{\`e}re et~al.(2023)Rozi{\`e}re, Gehring, Gloeckle, Sootla, Gat, Tan, Adi, Liu, Remez, Rapin, Kozhevnikov, Evtimov, Bitton, Bhatt, Ferrer, Grattafiori, Xiong, D'efossez, Copet, Azhar, Touvron, Martin, Usunier, Scialom, and Synnaeve]{Rozire2023CodeLO}
Baptiste Rozi{\`e}re, Jonas Gehring, Fabian Gloeckle, Sten Sootla, Itai Gat, Xiaoqing Tan, Yossi Adi, Jingyu Liu, Tal Remez, J{\'e}r{\'e}my Rapin, Artyom Kozhevnikov, I.~Evtimov, Joanna Bitton, Manish~P Bhatt, Cristian~Cant{\'o}n Ferrer, Aaron Grattafiori, Wenhan Xiong, Alexandre D'efossez, Jade Copet, Faisal Azhar, Hugo Touvron, Louis Martin, Nicolas Usunier, Thomas Scialom, and Gabriel Synnaeve.
\newblock Code llama: Open foundation models for code.
\newblock \emph{ArXiv}, abs/2308.12950, 2023.

\bibitem[Russakovsky et~al.(2014)Russakovsky, Deng, Su, Krause, Satheesh, Ma, Huang, Karpathy, Khosla, Bernstein, Berg, and Fei-Fei]{Russakovsky2014ImageNetLS}
Olga Russakovsky, Jia Deng, Hao Su, Jonathan Krause, Sanjeev Satheesh, Sean Ma, Zhiheng Huang, Andrej Karpathy, Aditya Khosla, Michael~S. Bernstein, Alexander~C. Berg, and Li~Fei-Fei.
\newblock Imagenet large scale visual recognition challenge.
\newblock \emph{International Journal of Computer Vision}, 115:\penalty0 211 -- 252, 2014.
\newblock URL \url{https://api.semanticscholar.org/CorpusID:2930547}.

\bibitem[Sagawa et~al.(2020)Sagawa, Koh, Hashimoto, and Liang]{Sagawa*2020Distributionally}
Shiori Sagawa, Pang~Wei Koh, Tatsunori~B. Hashimoto, and Percy Liang.
\newblock Distributionally robust neural networks.
\newblock In \emph{International Conference on Learning Representations}, 2020.

\bibitem[Santoro et~al.(2018)Santoro, Hill, Barrett, Morcos, and Lillicrap]{Santoro2018MeasuringAR}
Adam Santoro, Felix Hill, David G.~T. Barrett, Ari~S. Morcos, and Timothy~P. Lillicrap.
\newblock Measuring abstract reasoning in neural networks.
\newblock \emph{ArXiv}, abs/1807.04225, 2018.
\newblock URL \url{https://api.semanticscholar.org/CorpusID:49665167}.

\bibitem[Schick et~al.(2023)Schick, Dwivedi-Yu, Dessì, Raileanu, Lomeli, Zettlemoyer, Cancedda, and Scialom]{Schick2023Toolformer}
Timo Schick, Jane Dwivedi-Yu, Roberto Dessì, Roberta Raileanu, Maria Lomeli, Luke Zettlemoyer, Nicola Cancedda, and Thomas Scialom.
\newblock Toolformer: Language models can teach themselves to use tools.
\newblock In \emph{Advances in Neural Information Processing Systems (NeurIPS)}, volume~36, 2023.
\newblock URL \url{https://arxiv.org/abs/2302.04761}.

\bibitem[Setlur et~al.(2024)Setlur, Nagpal, Fisch, Geng, Eisenstein, Agarwal, Agarwal, Berant, and Kumar]{setlur2024rewarding}
Amrith Setlur, Chirag Nagpal, Adam Fisch, Xinyang Geng, Jacob Eisenstein, Rishabh Agarwal, Alekh Agarwal, Jonathan Berant, and Aviral Kumar.
\newblock Rewarding progress: Scaling automated process verifiers for llm reasoning.
\newblock \emph{arXiv preprint arXiv:2410.08146}, 2024.

\bibitem[Shi et~al.(2024)Shi, Lightman, Nye, Uesato, Bowman, and Kohli]{shi2024benchmarking}
Weijia Shi, Hunter Lightman, Maxwell Nye, Jonathan Uesato, Samuel~R Bowman, and Pushmeet Kohli.
\newblock Benchmarking process reward models for mathematical reasoning.
\newblock \emph{arXiv preprint arXiv:2401.08500}, 2024.

\bibitem[Shinn et~al.(2023)Shinn, Cassano, Labash, Gopinath, Narasimhan, and Yao]{Shinn2023ReflexionLA}
Noah Shinn, Federico Cassano, Beck Labash, Ashwin Gopinath, Karthik Narasimhan, and Shunyu Yao.
\newblock Reflexion: language agents with verbal reinforcement learning.
\newblock In \emph{Neural Information Processing Systems}, 2023.

\bibitem[Shojaee et~al.(2025)Shojaee, Mirzadeh, Alizadeh, Horton, Bengio, and Farajtabar]{shojaee2025illusionthinkingunderstandingstrengths}
Parshin Shojaee, Iman Mirzadeh, Keivan Alizadeh, Maxwell Horton, Samy Bengio, and Mehrdad Farajtabar.
\newblock The illusion of thinking: Understanding the strengths and limitations of reasoning models via the lens of problem complexity, 2025.
\newblock URL \url{https://arxiv.org/abs/2506.06941}.

\bibitem[Singhal et~al.(2023)Singhal, Tu, Gottweis, Sayres, Wulczyn, Hou, Clark, Liang, Corrado, Semturs, et~al.]{singhal2023towards}
Karan Singhal, Thang Tu, Jens Gottweis, Rory Sayres, Ellery Wulczyn, Luheng Hou, Chris Clark, Percy Liang, Greg Corrado, Christoph Semturs, et~al.
\newblock Towards expert-level medical question answering with large language models.
\newblock \emph{arXiv preprint arXiv:2305.09617}, 2023.

\bibitem[Singhal \& Shroff(2024)Singhal and Shroff]{singhal2024conceptsearch}
Kartik Singhal and Gautam Shroff.
\newblock Conceptsearch: Towards efficient program search using llms for abstraction and reasoning corpus (arc).
\newblock \emph{arXiv preprint arXiv:2412.07322}, 2024.
\newblock URL \url{https://arxiv.org/abs/2412.07322}.

\bibitem[Sinha et~al.(2025)Sinha, Arun, Goel, Staab, and Geiping]{sinha2025illusiondiminishingreturnsmeasuring}
Akshit Sinha, Arvindh Arun, Shashwat Goel, Steffen Staab, and Jonas Geiping.
\newblock The illusion of diminishing returns: Measuring long horizon execution in llms, 2025.
\newblock URL \url{https://arxiv.org/abs/2509.09677}.

\bibitem[Stanovich(2011)]{stanovich2011rationality}
Keith~E. Stanovich.
\newblock \emph{Rationality and the Reflective Mind}.
\newblock Oxford University Press, 2011.
\newblock ISBN 9780195341140.
\newblock \doi{10.1093/acprof:oso/9780195341140.001.0001}.

\bibitem[Stanovich \& West(2000)Stanovich and West]{Stanovich2000Individual}
Keith~E. Stanovich and Richard~F. West.
\newblock Individual differences in reasoning: Implications for the rationality debate?
\newblock \emph{Behavioral and Brain Sciences}, 23\penalty0 (5):\penalty0 645--665, 2000.

\bibitem[Sun et~al.(2025{\natexlab{a}})Sun, Li, Wang, Zhou, and Guo]{sun2025rlkd}
Jian Sun, Xinyun Li, Xia Wang, Shuyan Zhou, and Rui Guo.
\newblock Distilling the implicit multi-branch structure in llms’ reasoning via reinforcement learning.
\newblock \emph{arXiv preprint arXiv:2505.16142}, 2025{\natexlab{a}}.

\bibitem[Sun et~al.(2025{\natexlab{b}})Sun, Hsieh, Ladhak, Arakelyan, Serano, and Ginsburg]{sun2025l0reasoningbenchevaluatingprocedural}
Simeng Sun, Cheng-Ping Hsieh, Faisal Ladhak, Erik Arakelyan, Santiago~Akle Serano, and Boris Ginsburg.
\newblock L0-reasoning bench: Evaluating procedural correctness in language models via simple program execution, 2025{\natexlab{b}}.
\newblock URL \url{https://arxiv.org/abs/2503.22832}.

\bibitem[Sun et~al.(2025{\natexlab{c}})Sun, Hu, Zhou, Zheng, Hajishirzi, Dziri, and Song]{sun2025omegallmsreasonoutside}
Yiyou Sun, Shawn Hu, Georgia Zhou, Ken Zheng, Hannaneh Hajishirzi, Nouha Dziri, and Dawn Song.
\newblock Omega: Can llms reason outside the box in math? evaluating exploratory, compositional, and transformative generalization, 2025{\natexlab{c}}.
\newblock URL \url{https://arxiv.org/abs/2506.18880}.

\bibitem[Tan et~al.(2024)Tan, Shen, Chen, Courville, and Gan]{tan2024sparseuniversal}
Shawn Tan, Yikang Shen, Zhenfang Chen, Aaron Courville, and Chuang Gan.
\newblock Sparse universal transformer.
\newblock In \emph{International Conference on Learning Representations (ICLR) 2024}, 2024.

\bibitem[Taori et~al.(2020)Taori, Dave, Shankar, Carlini, Recht, and Schmidt]{taori2020measuring}
Rohan Taori, Achal Dave, Vaishaal Shankar, Nicholas Carlini, Benjamin Recht, and Ludwig Schmidt.
\newblock Measuring robustness to natural distribution shifts in image classification.
\newblock In \emph{Advances in Neural Information Processing Systems}, volume~33, pp.\  18583--18599. Curran Associates, Inc., 2020.
\newblock URL \url{https://proceedings.neurips.cc/paper/2020/hash/d8330f857a17c53d217014ee776bfd50-Abstract.html}.

\bibitem[Taori et~al.(2023)Taori, Gulrajani, Zhang, Liang, and Hashimoto]{taori2023llm}
Rohan Taori, Ishaan Gulrajani, Tianyi Zhang, Percy Liang, and Tatsunori~B. Hashimoto.
\newblock Llm evaluation beyond the imitation game.
\newblock \emph{arXiv preprint arXiv:2306.12822}, 2023.
\newblock URL \url{https://arxiv.org/abs/2306.12822}.

\bibitem[Ueda \& Washio(2021)Ueda and Washio]{ueda2021relationship}
Ryo Ueda and Koki Washio.
\newblock On the relationship between zipf’s law of abbreviation and interfering noise in emergent languages.
\newblock In \emph{Proceedings of the 59th Annual Meeting of the Association for Computational Linguistics and the 11th International Joint Conference on Natural Language Processing: Student Research Workshop}, pp.\  60--70, 2021.

\bibitem[Uesato et~al.(2022)Uesato, Kushman, Kumar, Song, Siegel, Wang, Creswell, Irving, and Higgins]{uesato2022solving}
Jonathan Uesato, Nate Kushman, Ramana Kumar, Francis Song, Noah Siegel, Lisa Wang, Antonia Creswell, Geoffrey Irving, and Irina Higgins.
\newblock Solving math word problems with process-and outcome-based feedback.
\newblock \emph{arXiv preprint arXiv:2211.14275}, 2022.

\bibitem[Vapnik(1998)]{Vapnik1998StatisticalLT}
Vladimir~Naumovich Vapnik.
\newblock Statistical learning theory.
\newblock 1998.

\bibitem[Vaswani et~al.(2017)Vaswani, Shazeer, Parmar, Uszkoreit, Jones, Gomez, Kaiser, and Polosukhin]{vaswani2017attention}
Ashish Vaswani, Noam Shazeer, Niki Parmar, Jakob Uszkoreit, Llion Jones, Aidan~N. Gomez, {\L}ukasz Kaiser, and Illia Polosukhin.
\newblock Attention is all you need.
\newblock In \emph{Advances in Neural Information Processing Systems}, volume~30, pp.\  5998--6008. Curran Associates, Inc., 2017.
\newblock \doi{10.5555/3295222.3295349}.
\newblock URL \url{https://proceedings.neurips.cc/paper/2017/hash/3f5ee243547dee91fbd053c1c4a845aa-Abstract.html}.

\bibitem[Veličković \& Blundell(2021)Veličković and Blundell]{Velickovic2021NeuralAlgorithmicReasoning}
Petar Veličković and Charles Blundell.
\newblock Neural algorithmic reasoning.
\newblock \emph{Patterns}, 2\penalty0 (7):\penalty0 100273, 2021.
\newblock \doi{10.1016/j.patter.2021.100273}.
\newblock URL \url{https://doi.org/10.1016/j.patter.2021.100273}.

\bibitem[Von Der~Malsburg(1986)]{von1986thinking}
Christoph Von Der~Malsburg.
\newblock Am i thinking assemblies?
\newblock In \emph{Brain Theory: Proceedings of the First Trieste Meeting on Brain Theory, October 1--4, 1984}, pp.\  161--176. Springer, 1986.

\bibitem[Wang et~al.(2025)Wang, Song, Tian, Yu, Mi, Duan, Tu, Su, and Yu]{wang2025fetch}
Ante Wang, Linfeng Song, Ye~Tian, Dian Yu, Haitao Mi, Xiangyu Duan, Zhaopeng Tu, Jinsong Su, and Dong Yu.
\newblock Don't get lost in the trees: Streamlining llm reasoning by overcoming tree search exploration pitfalls.
\newblock In \emph{Proceedings of the 63rd Annual Meeting of the Association for Computational Linguistics (ACL)}, pp.\  23946--23959, 2025.
\newblock \doi{10.18653/v1/2025.acl-long.1167}.
\newblock URL \url{https://aclanthology.org/2025.acl-long.1167.pdf}.

\bibitem[Wang et~al.(2023{\natexlab{a}})Wang, Xie, Jiang, Chen, and Ma]{wang2023makes}
Bailin Wang, Sang~Michael Xie, Yuxin Jiang, Shixiang Chen, and Tengyu Ma.
\newblock What makes reasoning hard in large language models?
\newblock \emph{arXiv preprint arXiv:2310.01798}, 2023{\natexlab{a}}.
\newblock URL \url{https://arxiv.org/abs/2310.01798}.

\bibitem[Wang et~al.(2024)Wang, Li, Shao, Xu, Dai, Li, Chen, Wu, and Sui]{wang2024math‌}
Peiyi Wang, Lei Li, Zhihong Shao, Runxin Xu, Damai Dai, Yifei Li, Deli Chen, Yu~Wu, and Zhifang Sui.
\newblock Math-shepherd: Verify and reinforce llms step-by-step without human annotations.
\newblock In \emph{Proceedings of the 62nd Annual Meeting of the Association for Computational Linguistics (Volume 1: Long Papers)}, pp.\  9426--9439, 2024.

\bibitem[Wang et~al.(2022)Wang, Wei, Schuurmans, Le, Chi, and Zhou]{Wang2022SelfConsistencyIC}
Xuezhi Wang, Jason Wei, Dale Schuurmans, Quoc Le, Ed~H. Chi, and Denny Zhou.
\newblock Self-consistency improves chain of thought reasoning in language models.
\newblock \emph{International Conference on Learning Representations}, 2022.

\bibitem[Wang et~al.(2023{\natexlab{b}})Wang, Wei, Schuurmans, Chi, Le, and Zhou]{wang2023process}
Xuezhi Wang, Jason Wei, Dale Schuurmans, Ed~Chi, Quoc Le, and Denny Zhou.
\newblock Process supervision improves mathematical reasoning of language models.
\newblock In \emph{NeurIPS}, 2023{\natexlab{b}}.

\bibitem[Wei et~al.(2022{\natexlab{a}})Wei, Tay, Bommasani, Zoph, Borgeaud, Yogatama, et~al.]{wei2022emergent}
Jason Wei, Yi~Tay, Rishi Bommasani, Barret Zoph, Sebastian Borgeaud, Dani Yogatama, et~al.
\newblock Emergent abilities of large language models.
\newblock \emph{Transactions on Machine Learning Research (TMLR)}, 2022{\natexlab{a}}.

\bibitem[Wei et~al.(2022{\natexlab{b}})Wei, Wang, Schuurmans, Bosma, Ichter, Xia, Chi, Le, and Zhou]{Wei2022ChainOT}
Jason Wei, Xuezhi Wang, Dale Schuurmans, Maarten Bosma, Brian Ichter, Fei Xia, Ed~H. Chi, Quoc~V. Le, and Denny Zhou.
\newblock Chain of thought prompting elicits reasoning in large language models.
\newblock In \emph{Advances in Neural Information Processing Systems (NeurIPS)}, volume~35, pp.\  24824--24837, 2022{\natexlab{b}}.
\newblock URL \url{https://proceedings.neurips.cc/paper_files/paper/2022/hash/b8d0553f5b7c2d301d9f0a1267a8d5c3-Abstract-Conference.html}.

\bibitem[Wei et~al.(2022{\natexlab{c}})Wei, Wang, Schuurmans, Bosma, Xia, Chi, Le, Zhou, et~al.]{wei2022chain}
Jason Wei, Xuezhi Wang, Dale Schuurmans, Maarten Bosma, Fei Xia, Ed~Chi, Quoc~V Le, Denny Zhou, et~al.
\newblock Chain-of-thought prompting elicits reasoning in large language models.
\newblock \emph{Advances in neural information processing systems}, 35:\penalty0 24824--24837, 2022{\natexlab{c}}.

\bibitem[Xia et~al.(2025)Xia, Wang, Liu, Li, Yu, Chen, McAuley, and Li]{xia2025beyond}
Yu~Xia, Rui Wang, Xu~Liu, Mingyan Li, Tong Yu, Xiang Chen, Julian McAuley, and Shuai Li.
\newblock Beyond chain-of-thought: A survey of chain-of-x paradigms for llms.
\newblock In \emph{Proceedings of the 31st International Conference on Computational Linguistics (COLING)}. Association for Computational Linguistics, 2025.
\newblock \doi{10.18653/v1/2025.coling-main.719}.
\newblock URL \url{https://aclanthology.org/2025.coling-main.719}.

\bibitem[Yao et~al.(2024)Yao, Yu, Zhao, Shafran, Griffiths, Cao, and Narasimhan]{yao2024tree}
Shunyu Yao, Dian Yu, Jeffrey Zhao, Izhak Shafran, Tom Griffiths, Yuan Cao, and Karthik Narasimhan.
\newblock Tree of thoughts: Deliberate problem solving with large language models.
\newblock \emph{Advances in Neural Information Processing Systems}, 36, 2024.

\bibitem[Ye \& Ng(2024)Ye and Ng]{Ye2024PreferenceGuidedRS}
Hai Ye and Hwee~Tou Ng.
\newblock Preference-guided reflective sampling for aligning language models.
\newblock In \emph{Conference on Empirical Methods in Natural Language Processing}, 2024.

\bibitem[Yu et~al.(2024)Yu, Peng, Vajipey, Cheng, Galley, Gao, and Yu]{yu2024rmcts}
Xiao Yu, Baolin Peng, Vineeth Vajipey, Hao Cheng, Michel Galley, Jianfeng Gao, and Zhou Yu.
\newblock Improving autonomous ai agents with reflective tree search and self-learning.
\newblock \emph{arXiv preprint arXiv:2410.02052}, 2024.
\newblock URL \url{https://arxiv.org/abs/2410.02052}.

\bibitem[Zan et~al.(2025)Zan, Wang, Zhou, and Schuurmans]{zan2025metacognition}
Yuhan Zan, Xuezhi Wang, Denny Zhou, and Dale Schuurmans.
\newblock Metacognition in large language models: Emergent self-evaluation and adaptive computation.
\newblock \emph{arXiv preprint arXiv:2504.08976}, 2025.

\bibitem[Zelikman et~al.(2022)Zelikman, Wu, Goodman, and Bowman]{zelikman2022star}
Eric Zelikman, Yuhuai Wu, Noah~D Goodman, and Samuel~R Bowman.
\newblock Self-taught reasoner: Bootstrapping reasoning with reasoning.
\newblock \emph{Advances in Neural Information Processing Systems (NeurIPS)}, 2022.

\bibitem[Zhang et~al.(2017)Zhang, Bengio, Hardt, Recht, and Vinyals]{zhang2017understanding}
Chiyuan Zhang, Samy Bengio, Moritz Hardt, Benjamin Recht, and Oriol Vinyals.
\newblock Understanding deep learning requires rethinking generalization.
\newblock In \emph{International Conference on Learning Representations}, 2017.
\newblock URL \url{https://openreview.net/forum?id=Sy8gdB9xx}.
\newblock ICLR 2017.

\bibitem[Zhang et~al.(2024)Zhang, Zhoubian, Hu, Yue, Dong, and Tang]{zhang2024restmcts}
Dan Zhang, Sining Zhoubian, Ziniu Hu, Yisong Yue, Yuxiao Dong, and Jie Tang.
\newblock Rest-mcts*: Llm self-training via process reward guided tree search.
\newblock \emph{arXiv preprint arXiv:2406.03816}, 2024.
\newblock URL \url{https://arxiv.org/abs/2406.03816}.

\bibitem[Zhou et~al.(2024)Zhou, Wang, Wei, Chi, Schuurmans, and Le]{zhou2024we}
Denny Zhou, Xuezhi Wang, Jason Wei, Ed~Chi, Dale Schuurmans, and Quoc Le.
\newblock We can’t always trust process supervision.
\newblock \emph{arXiv preprint arXiv:2402.04629}, 2024.

\bibitem[Zhou et~al.(2025{\natexlab{a}})Zhou, Wang, Wei, Chi, Le, and Schuurmans]{zhou2025verifree}
Denny Zhou, Xuezhi Wang, Jason Wei, Ed~Chi, Quoc Le, and Dale Schuurmans.
\newblock Verifying general reasoning without verifiers.
\newblock \emph{arXiv preprint arXiv:2505.21493}, 2025{\natexlab{a}}.

\bibitem[Zhou et~al.(2025{\natexlab{b}})Zhou, Liu, Chen, Tian, and Chen]{zhou2025gsminfinitellmsbehaveinfinitely}
Yang Zhou, Hongyi Liu, Zhuoming Chen, Yuandong Tian, and Beidi Chen.
\newblock Gsm-infinite: How do your llms behave over infinitely increasing context length and reasoning complexity?, 2025{\natexlab{b}}.
\newblock URL \url{https://arxiv.org/abs/2502.05252}.

\bibitem[Zhou et~al.(2025{\natexlab{c}})Zhou, Liu, Chen, Tian, and Chen]{zhou2025gsminfty}
Yang Zhou, Hongyi Liu, Zhuoming Chen, Yuandong Tian, and Beidi Chen.
\newblock {GSM}-\${\textbackslash}infty\$: How do your {LLM}s behave over infinitely increasing reasoning complexity and context length?
\newblock In \emph{Forty-second International Conference on Machine Learning}, 2025{\natexlab{c}}.

\end{thebibliography}
\bibliographystyle{tmlr}


\end{document}